\documentclass[10pt,twocolumn,letterpaper]{article}

\usepackage[pagenumbers]{iccv} 

\definecolor{iccvblue}{rgb}{0.21,0.49,0.74}
\usepackage[pagebackref,breaklinks,colorlinks,allcolors=iccvblue]{hyperref}
\usepackage{multirow}

\def\confName{ICCV}
\def\confYear{2025}

\def\eg{\textit{e.g.}}
\def\ie{\textit{i.e.}}

\usepackage[capitalize]{cleveref}
\crefname{section}{Sec.}{Secs.}
\crefname{table}{Tab.}{Tabs.}
\crefname{figure}{Fig.}{Figs.}
\newcommand{\methodname}{Free4D\xspace}

\title{\methodname: Tuning-free 4D Scene Generation with Spatial-Temporal Consistency}
\author{
Tianqi Liu$^{1,2,^*}$ \quad
Zihao Huang$^{1,2,^*}$ \quad
Zhaoxi Chen$^{2}$ \quad
Guangcong Wang$^{3}$ \quad \\
Shoukang Hu$^{2}$ \quad
Liao Shen$^{1,2}$ \quad
Huiqiang Sun$^{1,2}$ \quad
Zhiguo Cao$^{1}$ \quad
Wei Li$^{2,^\dagger}$ \quad
Ziwei Liu$^{2,^\dagger}$ 
\vspace{0.2em} \\
$^1$Huazhong University of Science and Technology, \\
$^2$S-Lab, Nanyang Technological University, \quad
$^3$Great Bay University \\
\href{https://free4d.github.io/}{https://free4d.github.io/}
}

\begin{document}

\twocolumn[{%
\renewcommand\twocolumn[1][]{#1}%
\maketitle
\begin{center}
    \centering
    \vspace{-20pt}
    \includegraphics[width=0.99\linewidth]{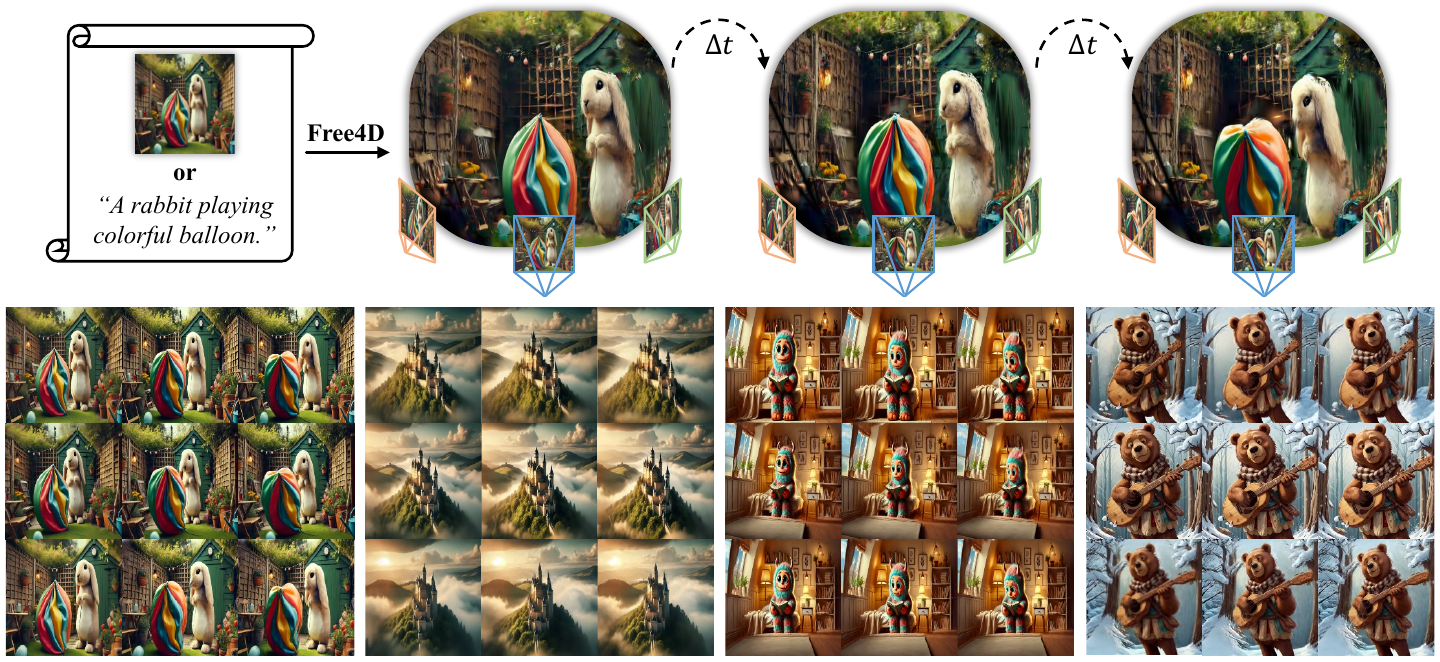}
    \vspace{-5pt}
    \captionof{figure}{\textbf{\methodname can generate diverse 4D scenes from single-image or textual input.} By enforcing spatial-temporal consistency in a tuning-free way, \methodname enables high-quality scene generation with explicit 4D controls.
    }
   \label{fig:teaser}
\end{center}%
}]

\renewcommand\thefootnote{} %
\footnotetext{$^*$Equal contribution. $^\dagger$Corresponding authors.}
\renewcommand\thefootnote{\arabic{footnote}} %

\begin{abstract}
We present \textbf{\methodname}, a novel tuning-free framework for 4D scene generation from a single image. 
Existing methods either focus on object-level generation, making scene-level generation infeasible, or rely on large-scale multi-view video datasets for expensive training, with limited generalization ability due to the scarcity of 4D scene data.
In contrast, our key insight is to distill pre-trained foundation models for consistent 4D scene representation, which offers promising advantages such as efficiency and generalizability.
\textbf{1)} To achieve this, we first animate the input image using image-to-video diffusion models followed by 4D geometric structure initialization.
\textbf{2)} To turn this coarse structure into spatial-temporal consistent multi-view videos, we design an adaptive guidance mechanism with a point-guided denoising strategy for spatial consistency and a novel latent replacement strategy for temporal coherence.
\textbf{3)} To lift these generated observations into consistent 4D representation, we propose a modulation-based refinement to mitigate inconsistencies while fully leveraging the generated information.
The resulting 4D representation enables real-time, controllable rendering, marking a significant advancement in single-image-based 4D scene generation.
\end{abstract}

\vspace{-20pt}

\section{Introduction}
\label{sec:intro}
Creating a dynamic 3D environment that closely mirrors the real world is crucial for achieving realistic and immersive digital experiences, a key objective in fields such as film production, video games, and augmented reality.
However, captured images provide only limited snapshots of a scene. Generating dynamic 3D scenes from such limited observations, particularly from a single image, remains a significant challenge and an open research problem.

Existing 4D generation methods~\cite{dream-in-4d,4dfy,dreamgaussian4d,l4gm} primarily focus on objects, often neglecting background generation and its dynamics. 
Recently, several studies~\cite{dimensionX,cat4d,genXD,4real} have explored 4D scene generation for real-world applications. Compared to single-object generation, scene-level 4D generation presents greater challenges, requiring handling complex geometries, spatial relationships, and dynamic interactions.
One line of research relies on fine-tuned video diffusion models to get temporally varying multi-view data for 4D representation fitting.
By decoupling spatial and temporal dimensions~\cite{dimensionX} or enforcing consistency alternately~\cite{cat4d}, these methods aim to generate coherent multi-view videos, which can be optimized to construct a 4D representation of the scene. 
Some works~\cite{genXD} further introduce data curation pipelines to generate 4D scene data for training models. 
However, the performance of these methods is highly dependent on the quality and scale of generated 4D scene data. 
Moreover, they require substantial data and computational resources to fine-tune large video diffusion models. 
Another line of research leverages priors from existing generative models to optimize a 4D representation, avoiding the high costs associated with fine-tuning diffusion models. 
%
%
A pioneering work~\cite{4real} achieves text-to-4D scene generation by distilling priors from a closed-source video diffusion model~\cite{snap} using score distillation sampling (SDS)~\cite{dreamfusion}, producing impressive results. However, it inherits common limitations of SDS, including lengthy optimization, oversaturated colors~\cite{prolificdreamer, wei2024adversarial}, and limited diversity~\cite{prolificdreamer, wei2024adversarial} in the generated outputs.

To overcome these limitations, we propose \textbf{\methodname}, a 4D scene generation method from a single image with explicit spatial-temporal consistency in a data-efficient and tuning-free manner.
To obtain a 4D representation from a single image, a straightforward solution would be generating a multi-view video from the given image and then optimize a 4D representation based on it. However, this approach presents two key challenges:
\textbf{1)} How to generate a multi-view video from a single image while ensuring high spatial-temporal consistency.  
\textbf{2)} How to effectively optimize a coherent 4D representation despite inevitable minor inconsistencies in the generated multi-view video.

To address the \textbf{first} challenge, we adopt the dynamic reconstruction method~\cite{monst3r}, enhanced by progressive background point cloud aggregation strategy. This approach enables accurate initialization of a coherent 4D geometric structure, thus ensuring geometric alignment for subsequent generation.
Subsequently, guided by the established 4D structure, we employ a point-conditioned diffusion model~\cite{viewcrafter} to generate a multi-view video. To enhance spatial consistency, we introduce two strategies: an \textit{adaptive classifier-free guidance (CFG)} approach, designed to maintain consistent appearance across different viewpoints, and \textit{point cloud guided denoising}, aimed at reducing unintended motions of dynamic subjects in the synthesized views.
Although these strategies notably improve spatial consistency, significant temporal inconsistencies remain, primarily due to the generative model's tendency to produce inconsistent content in occluded or missing regions over time. To mitigate this, we propose \textit{reference latent replacement}, a technique that substantially enhances temporal coherence, ensuring smoother and more consistent video content over time.

With these advancements, the generated multi-view video achieves near-complete spatial-temporal consistency. However, subtle inconsistencies persist, posing challenges in constructing a fully coherent 4D representation.
To overcome this \textbf{second} challenge, we introduce an effective optimization strategy designed to seamlessly integrate multi-view videos into a unified 4D representation. Our approach begins by constructing a coarse 4D representation, utilizing only those images that share the same timestamp or viewpoint as the input image. To further refine this representation, we incorporate a \textit{modulation-based refinement}, leveraging additional generated images while effectively suppressing inconsistencies. 
The resulting 4D representation enables real-time, controllable spatial-temporal rendering, ensuring both fidelity and coherence across views and time.

Our main contributions can be summarized as follows:
\begin{itemize}[leftmargin=*]
    \item We present \methodname, the first tuning-free pipeline for 4D scene generation from a single image, delivering photo-realistic appearances and realistic motions.
    \item We employ a dynamic point-conditioned multi-view video generation approach, integrating carefully designed techniques to enhance spatial-temporal consistency.
    \item We introduce a coarse-to-fine training strategy combined with a modulation-based refinement to effectively integrate the generated information while reducing inconsistencies, yielding a consistent 4D representation.
\end{itemize}
\section{Related Work}
\label{sec:related work}
\noindent \textbf{Video Generation.} Pioneering studies achieve dynamic video generation with VAEs~\cite{DBLP:conf/icml/DentonF18, DBLP:journals/corr/RanzatoSBMCC14, DBLP:conf/icml/KalchbrennerOSD17, babaeizadeh2021fitvid, DBLP:conf/iclr/KumarBEFLDK20, DBLP:journals/corr/MathieuCL15} or GANs~\cite{DBLP:conf/nips/VondrickPT16, clark2019adversarial, tulyakov2018mocogan, DBLP:conf/wacv/WangBBD20}, while facing limitations in temporal stability and output resolution.
Subsequent methods revolutionize this field by extending image diffusion models with 3D UNet architecture~\cite{VDM} or auto-regressive manners~\cite{harvey2022flexible}.
Following advancements introduced controllable frameworks that enabled synthesis guided by texts~\cite{cogvideo, make-a-video, DBLP:conf/cvpr/HuLC22, DBLP:journals/corr/abs-2210-02303}, images~\cite{DBLP:conf/iclr/BabaeizadehFECL18, DBLP:conf/eccv/LiFYWLY18, pan2019video, blattmann2021understanding} or viewpoints~\cite{SynCamMaster, Collaborative-Video-Diffusion} through conditional denoising techniques.
Recent advancements focus on enhancing the realism and detail of generated videos~\cite{videocrafter2, videofusion, Lavie, DBLP:conf/cvpr/BlattmannRLD0FK23, DBLP:conf/iccv/GeNLPTC0H0B23, DBLP:conf/iclr/0002YRL00AL024, DBLP:journals/corr/abs-2211-13221}, capturing intricate details and natural movements, making generated videos more lifelike and engaging.
Despite these successes, video generation inherently lacks explicit 3D scene structures or support for viewpoint manipulation—critical gaps for 3D / 4D spatial-temporal modeling tasks.

\noindent \textbf{3D~/~4D Generation.}
Early explorations in 3D generation focused on static objects through point clouds~\cite{DBLP:conf/iclr/AchlioptasDMG18, get3d, DBLP:journals/corr/abs-2409-12957} or implicit neural representations~\cite{DBLP:conf/cvpr/ParkFSNL19, DBLP:conf/iccv/ChouBH23}. Extending to 3D scene generation, SceneDreamer~\cite{DBLP:journals/pami/ChenWL23} leveraged neural radiance fields (NeRF)~\cite{nerf} for unbounded outdoor generation, while InfiniCity~\cite{DBLP:conf/iccv/LinLMCS0T23} proposed scalable pipelines for photorealistic urban scenes. However, these methods primarily focus on static scene generation and lack support for dynamic spatial-temporal modeling.
Subsequent methods~\cite{DBLP:journals/corr/abs-2412-20422, eg4d, 4dgen, pan2024efficient4d, make-a-video4d, dream-in-4d, animate124, l4gm, dreamgaussian4d} further introduce temporal deformations, enabling controllable 4D generation of single objects. Recent approaches~\cite{dimensionX, genXD} attempted to unify dynamic objects and environmental interactions through space-time neural representations. 
However, these methods heavily depend on large-scale, high-quality 4D training data, which are labor-intensive to acquire and often restrict real-world applicability. 
The method most closely related to ours is 4Real~\cite{4real}, which only supports text conditions and relatively low resolution.

\noindent \textbf{4D Reconstruction.}
Neural Radiance Fields (NeRF)~\cite{nerf, d-nerf} represents 3D scenes via implicit neural representations, while 3D Gaussian Splatting~\cite{3dgs, 2dgs} later introduces explicit, real-time 3D primitives. 
Extending these to 4D, recent advances propose dynamic 3D Gaussians, where Gaussian attributes evolve via deformation fields~\cite{4dgs, DBLP:conf/cvpr/YangGZJ0024}. Further innovations~\cite{dy3dgs, dymarbles} optimize spatiotemporal Gaussian kernels directly from RGB inputs. While these methods achieve photo-realistic 4D reconstruction, they remain tightly coupled with high-fidelity 4D training data, which limits scalability for real-world dynamic scenes.

\section{Preliminaries}
\label{sec:preliminary}
\noindent \textbf{Latent Diffusion Model (LDM)}~\cite{ldm} is a computationally efficient variant of diffusion models that both the forward and the reverse process are performed in the latent space.
Given an image $x_0$, it is first encoded into a latent representation $z_0 = \mathcal{E}(x_0)$ using a VAE encoder $\mathcal{E}$. The forward process progressively adds noise $\epsilon$, formulated as:
\begin{equation}
    z_i =  \sqrt{1-\beta _i} z_{i-1} + \sqrt{\beta _i} \epsilon, 
\label{eq:forward_process}
\end{equation}
where $\beta_i \in (0,1)$ represents the noise schedule at time step $i$.
The cumulative noise follows the closed-form expression:
\begin{equation}
    z_i = \sqrt{\bar{\alpha}_i}  z_{0} + \sqrt{1-\bar{\alpha}_i}\epsilon, 
\label{eq:forward_closed_form}
\end{equation}
where $\bar{\alpha}_i =  {\textstyle \prod_{1}^{i}} (1-\beta _i)$.
The reverse process removes noise from latent. We adopt DDIM~\cite{ddim}, given by:
\begin{equation}
z_{i-1} = \sqrt{\bar{\alpha}_{i-1}} z_{0 \gets i} + \sqrt{1 - \bar{\alpha}_{i-1}} \epsilon_{\theta}(z_i, i),
\label{eq:ddim}
\end{equation}
where ${\epsilon}_{\theta}(z_i, i)$ denotes the predicted noises.
Combining \cref{eq:forward_closed_form} and \cref{eq:ddim}, the denoising process are rewritten as:
\begin{equation}
z_{i-1}=a_i z_i+b_i  {z}_{0 \gets i},
\label{eq:ddim_2}
\end{equation}
where
$
a_i=\sqrt{\frac{1-\bar{\alpha}_{i-1}}{1-\bar{\alpha}_i}},
b_i=\sqrt{\bar{\alpha}_{i-1}}-\sqrt{\bar{\alpha}_i}  a_i
$,
and
$
z_{0 \gets i} = {(z_i - \sqrt{1 - \bar{\alpha}_i} {\epsilon}_{\theta}(z_i, i))}/{\sqrt{\bar{\alpha}_i}}
$.
\cref{eq:ddim_2} indicates that the denoising direction is determined by ${z}_{0 \gets i}$.
Finally, the generated image is obtained via the VAE decoder: $\hat{x} = \mathcal{D}(z_0)$.
\section{Free4D}
\label{sec:method}

\begin{figure*}[!t]
  \centering
    \includegraphics[width=0.99\linewidth]{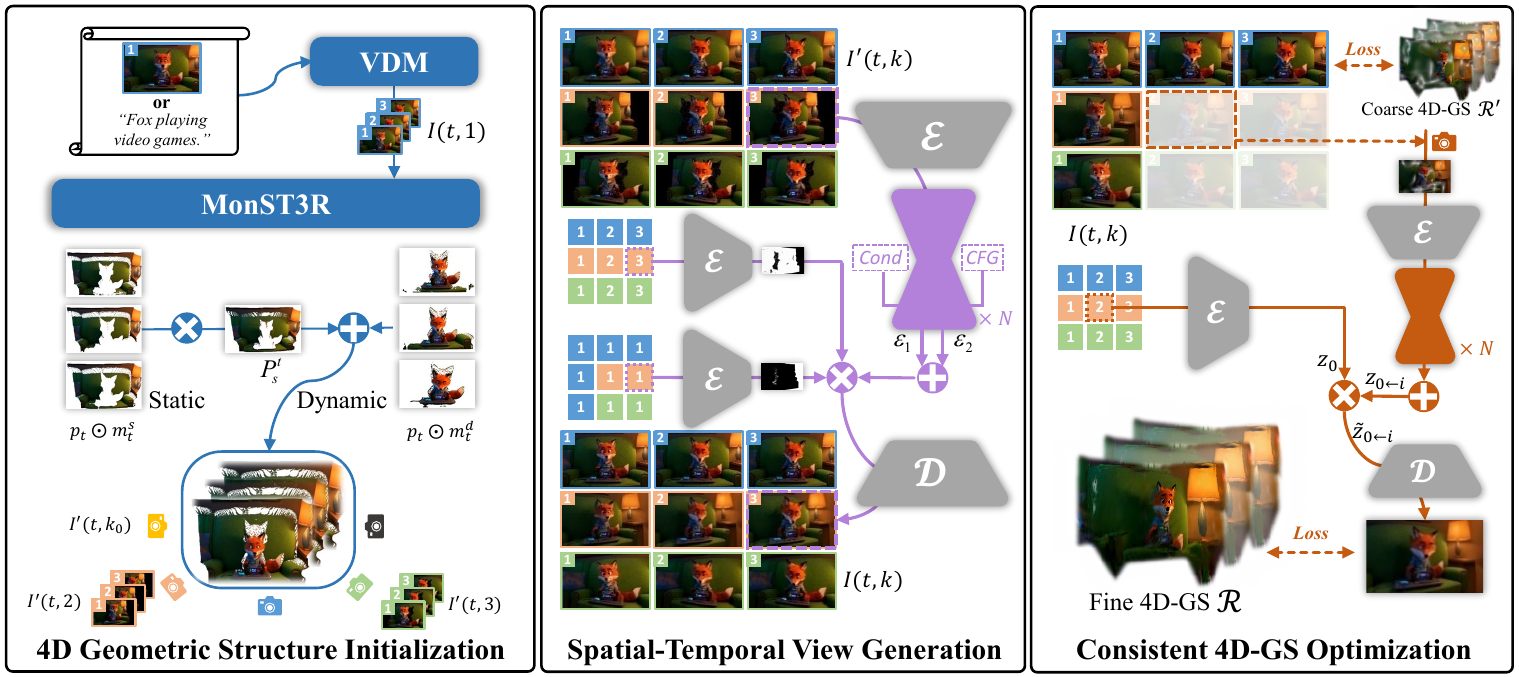}
    \vspace{-5pt}
    \caption{\textbf{Overview of \methodname.} Given an input image or text prompt, we first generate a dynamic video $\mathcal{V}=\{I(t,1)\}_{t=1}^{T}$ using an off-the-shelf video generation model~\cite{kling}. Then, we employ MonST3R~\cite{monst3r} with a progressive static point cloud aggregation strategy for dynamic reconstruction, obtaining a 4D geometric structure. Next, guided by this structure, we render a coarse multi-view video $\mathcal{S}^{\prime}=\{\{I^{\prime}(t,k)\}_{t=1}^{T}\}_{k=1}^{K}$ along a predefined camera trajectory and refine it into $\mathcal{S}=\{\{I(t,k)\}_{t=1}^{T}\}_{k=1}^{K}$ using ViewCrafter~\cite{viewcrafter}. To ensure spatial-temporal consistency, we introduce Adaptive Classifer-Free Guidance (CFG) and Point Cloud Guided Denoising for spatial coherence, along with Reference Latent Replacement for temporal coherence. Finally, we propose an efficient training strategy with a Modulation-Based Refinement to lift the generated multi-view video $\mathcal{S}$ into a consistent 4D representation $\mathcal{R}$.}
    \label{fig:pipeline}
    \vspace{-10pt}
\end{figure*}

Given a single scene image $I$, we aim to derive feasible 4D Gaussian representations with \textit{spatial-temporal consistency} in a \textit{tuning-free} approach, which enables the synthesis of high-fidelity novel views in free viewpoints \textit{at minimal cost}. 
To achieve this, we start by converting the input image into a video $\mathcal{V}$ using an off-the-shelf image-to-video generator and initializing the associated 4D geometric structures $\mathcal{P}$.
Then, we produce spatial-temporal consistent multi-view videos $\mathcal{S}$ using a point-conditioned diffusion model with guidance from the obtained 4D geometric structures. The final 4D representation $\mathcal{R}$ is optimized via a proposed training strategy to further improve spatial-temporal consistency.

\subsection{4D Geometric Structure Initialization}
\label{sub:dpc}

With an image-to-video generative model~\cite{kling}, we first animate the input scene image \( I \) into a reference single-view video $\mathcal{V} = \{I(t, 1)\}_{t=1}^{T}$. Here, an image at time \( t \) and viewpoint \( k \) in the multi-view video is denoted as \( I(t, k) \).
To initialize 4D geometric structures from the reference single-view video $\mathcal{V}$, we employ point clouds $\mathcal{P}=\{P_t\}_{t=1}^{T}$ as explicit representations. This is crucial for enhancing geometric consistency and camera control capabilities for 4D scene generation.
Specifically, we apply a dynamic scene reconstruction method MonST3R~\cite{monst3r}, which takes the reference video $\mathcal{V}$ as inputs and produces world-coordinate pointmaps $\{p_t\}_{t=1}^{T}$.  Simultaneously, per-frame static masks $\{m_t^s\}_{t=1}^{T}$ for distinguishing invariant regions within each image are estimated. To effectively integrate geometry information within the reference video, we convert initial pointmaps into well-organized point clouds. Considering that directly using naive pointmaps might neglect some cross-frame geometry information caused by occlusions, we decompose pointmaps into static and dynamic components using static masks $\{m_t^s\}_{t=1}^{T}$.
We aim to keep static components consistent across frames while dynamic components vary over time in per-frame point clouds. One might concatenate all static regions in all frames straightforwardly, but this leads to redundant points and inefficiency in the subsequent rendering process. Therefore, we propose a progressive strategy that can effectively aggregate static components. We initialize point clouds from static regions in the first frame, $P_{1}^{s} = p_{1} \odot m^{s}_{1}$, given that the first frame (\ie, the input scene image \( I \)) contains the highest confidence and quality. We progressively update $P_{1}^{s}$  by propagating static regions from subsequent frames while avoiding redundancy: 
\begin{equation}
P_t^s = P_{t-1}^s \cup (p_t \odot \hat{m}_t^s),
\end{equation}
where $\hat{m}_{t}^s = m_{t}^s \cap (1- \bigcup_{i=1}^{t-1} m_{i}^s)$
and \( \odot \) denotes element-wisely indexing none-zero values.
This ensures a compact yet complete representation of the static point cloud while maintaining alignment and consistency across frames. 
The dynamic components in each frame are kept in their corresponding pointmaps. Thus, the point cloud $\{P_t\}_{t=1}^{T}$ is given by $P_{t} = P^s_T \cup ( p_{t} \odot m^d_{t})$, where $m^d_{t} = 1 - m^s_{t}$.

\subsection{Spatial-Temporal View Generation}
\label{sub:view generation}

Due to the scarcity of 4D data, there is no available off-the-shelf multi-view video generator. Therefore, we propose a \textit{tuning-free} approach to generate multi-view videos $\mathcal{S}$ that maintain robust spatial-temporal consistency, with the guidance of the obtained 4D geometry structures, \ie, point clouds $\mathcal{P}=\{P_t\}_{t=1}^{T}$. We consider rendering the point clouds from different camera poses to create a sequence of coarse multi-view videos $\mathcal{S}^{\prime} = \{\{I^{\prime}(t, k)\}_{t=1}^{T}\}_{k=1}^{K}$ together with visibility masks $\mathcal{M} =\{\{M(t, k)\}_{t=1}^{T}\}_{k=1}^{K}$, with navigation of a user-defined camera trajectory ($K$ camera poses). 
Despite capturing geometric structures and view relationships effectively, the coarse multi-view videos still face challenges such as occlusions, missing regions, and diminished visual fidelity.  

To mitigate these issues, we employ ViewCrafter~\cite{viewcrafter}, a point-conditioned diffusion model, to refine the coarse multi-view videos. However, simply using ViewCrafter cannot guarantee strong spatial-temporal consistency in refined multi-view videos. There are two main problems: 1) From a spatial consistency perspective, it fails to maintain uniform color tones across frames and introduces unexpected motion artifacts in scenes with highly dynamic content.
2) In terms of temporal consistency, it generates noticeable discrepancies between consecutive frames, leading to temporal flickering and a lack of smooth transitions. We tackle these issues in the following ways.

\noindent\textbf{Geometry-informed Adaptive Denoising.}
In ViewCrafter, using the naive classifier-free guidance (CFG)~\cite{cfg} tends to accumulate numerical errors and cause over-saturation problems~\cite{mokady2023null}. Disabling CFG by setting the guidance scale to $1$ would generate low-quality results or even failure to complete in some scenarios~(\cref{subsec:ablations}). We propose an \textbf{\textit{Adaptive CFG}} strategy by deactivating CFG in regions where the point cloud rendering is visible (\ie, $M(t, k)=1$) and compute the predicted noise as follows:
\begin{equation}
\epsilon_{\text{1}} = \epsilon_{\theta} (z_i, c),
\label{eq:cond}
\end{equation}
where $z_i$ represents the latent of a specific image $I{(t,k)}$ at denoising timestep $i$, $\epsilon_{\theta}$ denotes the denoising network, and $c$ is the condition, 
specifically the image $I{(t,1)}$ and a default prompt text used by~\cite{viewcrafter}. On the contrary, for occluded or missing regions (\ie, $M(t, k)=0$), we enable CFG and compute the noise as:
\begin{equation}
\epsilon_{\text{2}} = \epsilon_{\theta} (z_i) + s \cdot (\epsilon_{\theta} (z_i, c) - \epsilon_{\theta} (z_i)),
\label{eq:cfg}
\end{equation}
where $s$ is the guidance scale. Thus, the final estimated noise is obtained by a noise fusion at each denoising step :
\begin{equation}
\epsilon= M(t, k) \cdot \epsilon_{\text{1}} + (1-M(t, k)) \cdot \epsilon_{\text{2}}.
\label{eq:ada_cfg}
\end{equation}
Moreover, we propose \textbf{\textit{Point Cloud Guided Denoising}} by leveraging the coarse multi-view video $\mathcal{S}^{\prime}$ to guide the early denoising process. 
Specifically,  we encode the specific image $I^{\prime}(t,k)$ in $\mathcal{S}^{\prime}$ into latent representations:
\begin{equation}
z^{\prime}_0 = \mathcal{E} (I^{\prime}(t,k)),
\label{eq:vae}
\end{equation}
At an early denoising timestep $i$, we first apply the forward diffusion process to $z^{\prime}_0$ to obtain noisy latent $z^{\prime}_i$ following~\cref{eq:forward_closed_form}. We then fuse $z^{\prime}_i$ with the model-predicted latent $z_i$ based on the point cloud rendered mask $m=M(t, k)$, which is given by:
\begin{equation}
\hat{z}_i = m \cdot z^{\prime}_i + (1 - m) \cdot z_i. 
\label{eq:pgd}
\end{equation}
By employing this adaptive approach that leverages information from point cloud renders $\mathcal{S}^{\prime}$ to guide the denoising process, we effectively mitigate color inconsistencies, reduce unexpected dynamic motion, and enhance spatial consistency across views.

\noindent\textbf{Consistent Temporal Latent  Replacement.} We further refine the point cloud renders $\mathcal{S}^{\prime}$ at different timestamps to enhance temporal consistency in multi-view generation by proposing a \textbf{\textit{Reference Latent Replacement}} strategy. Specifically, for a specific timestamp $t_j>1$, we use the multi-view images $\{I{(1,k)}\}_{k=1}^{K}$ generated from the first frame as a reference. 
For simplicity, we illustrate the following process using a specific \( k_j \in [1, K] \). For generating image $I{(t_j,k_j)}$, we use the first-timestamp $I{(1,k_j)}$ as a reference. In regions where both $I{(t_j,k_j)}$ and $I{(1,k_j)}$ require completion, \ie, $M(t_j, k_j)=0$ and $M(1, k_j)=0$, we replace the latent in these areas with those from the reference. 
Specifically, we first encode $I{(1,k_j)}$ into a latent as:
\begin{equation}
z^{ref}_0 = \mathcal{E} (I{(1,k_j)}). 
\label{eq:vae_ref}
\end{equation}
At a denoising timestep $i$, we first apply the forward diffusion process to $z^{ref}_0$ to obtain noisy latent $z^{ref}_i$ following~\cref{eq:forward_closed_form} and  predict the latent $z_i$ for $I{(t_j,k_j)}$. We then fuse $z^{ref}_i$ with $z_i$ based on the co-visible mask:
\begin{equation}
\hat{m}=(1-M(t_j, k_j)) \cdot (1-M(1, k_j)).
\label{eq:fuse}
\end{equation}
The replaced latent $\hat{z}_i$ is given by: 
\begin{equation}
\hat{z}_i = \hat{m} \cdot z^{ref}_i + (1 - \hat{m}) \cdot z_i. 
\label{eq:replaced}
\end{equation}
This approach preserves consistency in the generated content over time, effectively reducing discrepancies and producing multi-view videos that achieve nearly-consistency in both temporal and spatial dimensions.

\subsection{Consistent 4D-GS Optimization}
\label{sub:train 4D-GS}

Given generated spatial-temporal consistent multi-view videos, we optimize the corresponding 4D-GS representations. Due to the high-dynamic property of generated multi-view videos, directly using a standard training pipeline with the multi-view video as supervision would cause misalignment and inconsistency in the final 4D-GS. Therefore, we propose an effective training strategy to integrate the information from the generated multi-view videos for consistent 4D-GS optimization. Our key insight is that the consistency between the reference video $\{I{(t,1)}\}_{t=1}^{T}$ at $k=1$ and the generated multi-view images $\{I{(1,k)}\}_{k=1}^{K}$ at the first timestamp $t=1$ is relatively high, as both are constrained by the input image. Thus, we first utilize these views to train a coarse 4D-GS $\mathcal{R}^{\prime}$. Then, we incorporate the missing information from the rest of the multi-views to obtain a refined 4D-GS $\mathcal{R}$.
However, it is difficult to extract useful information while preventing the propagation of inconsistencies into the 4D-GS representation. Instead of using generated images for pixel-level supervision directly,  we integrate generated information into the 4D representation at a higher level.
Specifically, we first render the coarse 4D-GS at specific \( t_j \) and \( k_j \) to obtain a rendered image \( I^r \).
We then apply the forward process of the diffusion model by adding noise to obtain the noisy renderings \( z^r_{\bar{T}} \), where \( \bar{T} \) is a predefined timestep.
During the denoising stage, inspired by~\cite{vistadream}, we introduce a \textbf{\textit{Modulation-Based Refinement}} for effective enhancement.
We use the generated image $I(t_j,k_j)$ as a modulation signal at each timestep, guiding the denoising process toward the desired generated context. In~\cref{eq:ddim_2}, since \( {z}^{r}_{0 \gets i}\) (at the denoising timestep \( i \) ) estimates the noise-free latent of the rendered image and dictates the denoising direction, we propose integrating the information from the generated image into this process to adjust the denoising direction. The generated image is first encoded into a latent as $z_{0}=\mathcal{E} (I(t_j,k_j))$, and the adjusted process is given by:
\begin{equation}
    \tilde{z}_{0 \gets i} = w_i \gamma_i z_{0} + (1-w_i) z_{0 \gets i},
    \label{eq:w}
\end{equation}
where \( \gamma_i = \frac{\operatorname{std}(z_{0 \gets i})}{\operatorname{std}(z_0)}\) serves as a scaling factor to mitigate over-exposure~\citep{stdalign,vistadream}, while \( w_i \) is a predefined weight that regulates the influence of the generated image on the denoising process. We replace the original \( z_{0 \gets i} \) with adjusted \( \tilde{z}_{0 \gets i} \) for the subsequent denoising process.  This adjustment is applied at each denoising step to obtain the enhanced renderings, denoted as \( \tilde{I^r} \), which are used to improve the coarse 4D-GS and enhance both rendering quality and consistency.

\noindent \textbf{Loss Function.}  
For the first timestamp (\( t=1 \)) or first viewpoint (\( k=1 \)), we use L1 loss:
\begin{equation}
L_{l1} = \| I(t, k) - I^{r}(t, k) \|_1
\end{equation}
where \( I(t, k) \) and \( I^r(t, k) \) represent the generated image and rendered image by 4D-GS, respectively.  
For other images ($t>1, k>1$), we use LPIPS loss~\cite{lpips}, as:
\begin{equation}
L_{lpips} = \text{LPIPS}(\tilde{I^r}(t,k), I^{r}(t, k))
\end{equation}
where $\tilde{I^r}(t,k)$ is the refined generated image.
For the coarse stage, the total loss is $L=L_{l1}$ while for the fine stage, the total loss is $L=L_{l1}+\lambda L_{lpips}$.

\begin{figure*}[!t]
  \centering
    \includegraphics[width=0.99\linewidth]{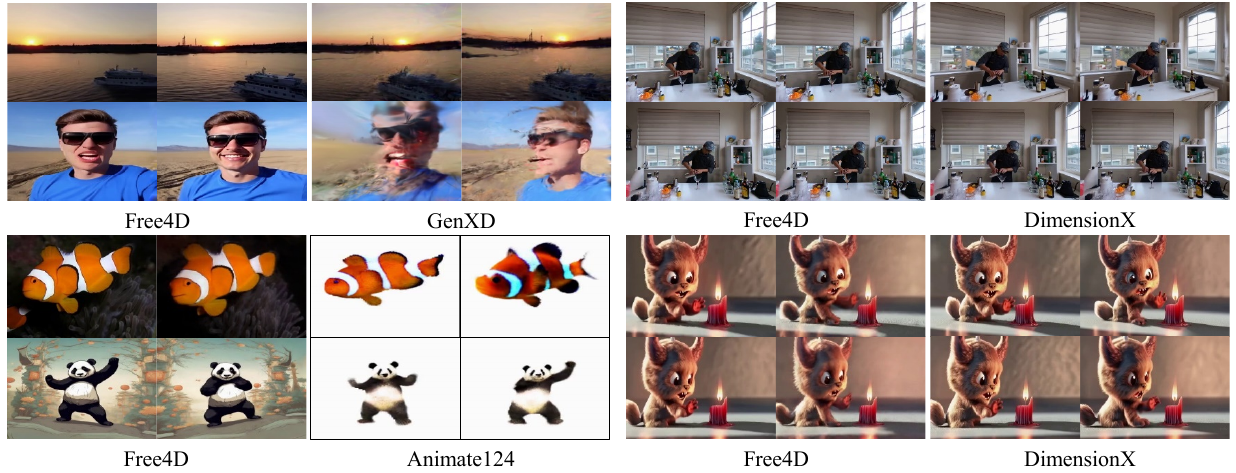}
    \vspace{-5pt}
    \caption{\textbf{Qualitative comparisons of image-to-4D.}
    We present the results using the same single-image prompts.
    }
    \vspace{-5pt}
    \label{fig:comparison-img}
\end{figure*}

\section{Experiments}
\label{sec:experiments}

\begin{figure*}[!t]
  \centering
    \includegraphics[width=0.99\linewidth]{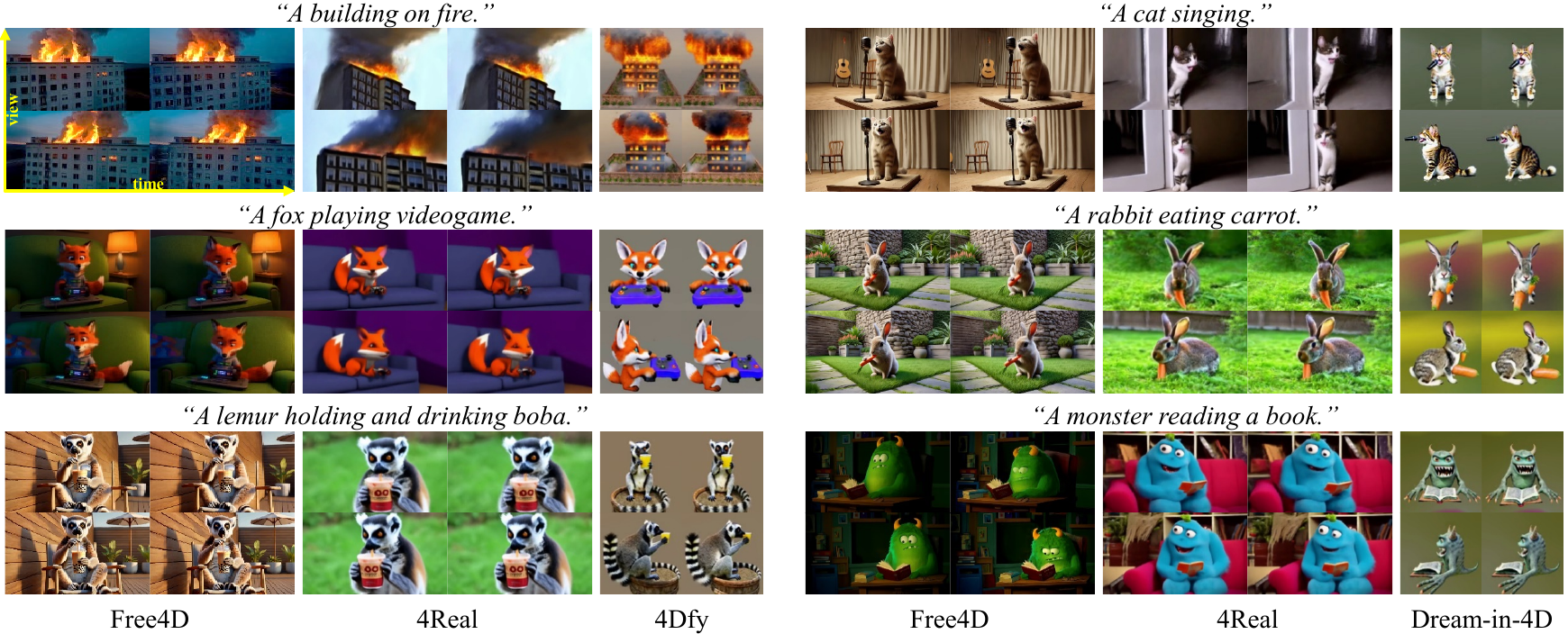}
    \vspace{-5pt}
    \caption{\textbf{Qualitative comparisons of text-to-4D.} We show the results based on the same text prompts.
    }
    \vspace{-10pt}
    \label{fig:comparison-text}
\end{figure*}

\noindent \textbf{Baselines.}
Our baselines fall into two categories: text-to-4D and image-to-4D methods.
For text-to-4D, we compare our approach with 4Real~\cite{4real}, a state-of-the-art text-to-4D scene generation method. We also include two widely used object-centric 4D generation methods: 4Dfy~\cite{4dfy} and Dream-in-4D~\cite{dream-in-4d}. 
For image-to-4D, we compare our approach with two state-of-the-art generative models: DimensionX~\cite{dimensionX} and GenXD~\cite{genXD}, both trained on large-scale datasets while our \methodname operates without tuning. We also include Animate124~\cite{animate124}, a state-of-the-art object-centric tuning-free method based on SDS~\cite{dreamfusion}.
Specifically, we use the same text or single-image prompts for generation.
Since the official implementations of 4Real~\cite{4real}, DimensionX~\cite{dimensionX}, and GenXD~\cite{genXD} have not been released, we report the results available from their respective project pages.
GenXD only releases generated videos instead of videos rendered from the 4D representation. Therefore, we use a monocular reconstruction algorithm~\cite{yang2024deformable} to reconstruct the 4D representation and render it for comparison.

\begin{figure*}[!t]
  \centering
    \includegraphics[width=0.99\linewidth]{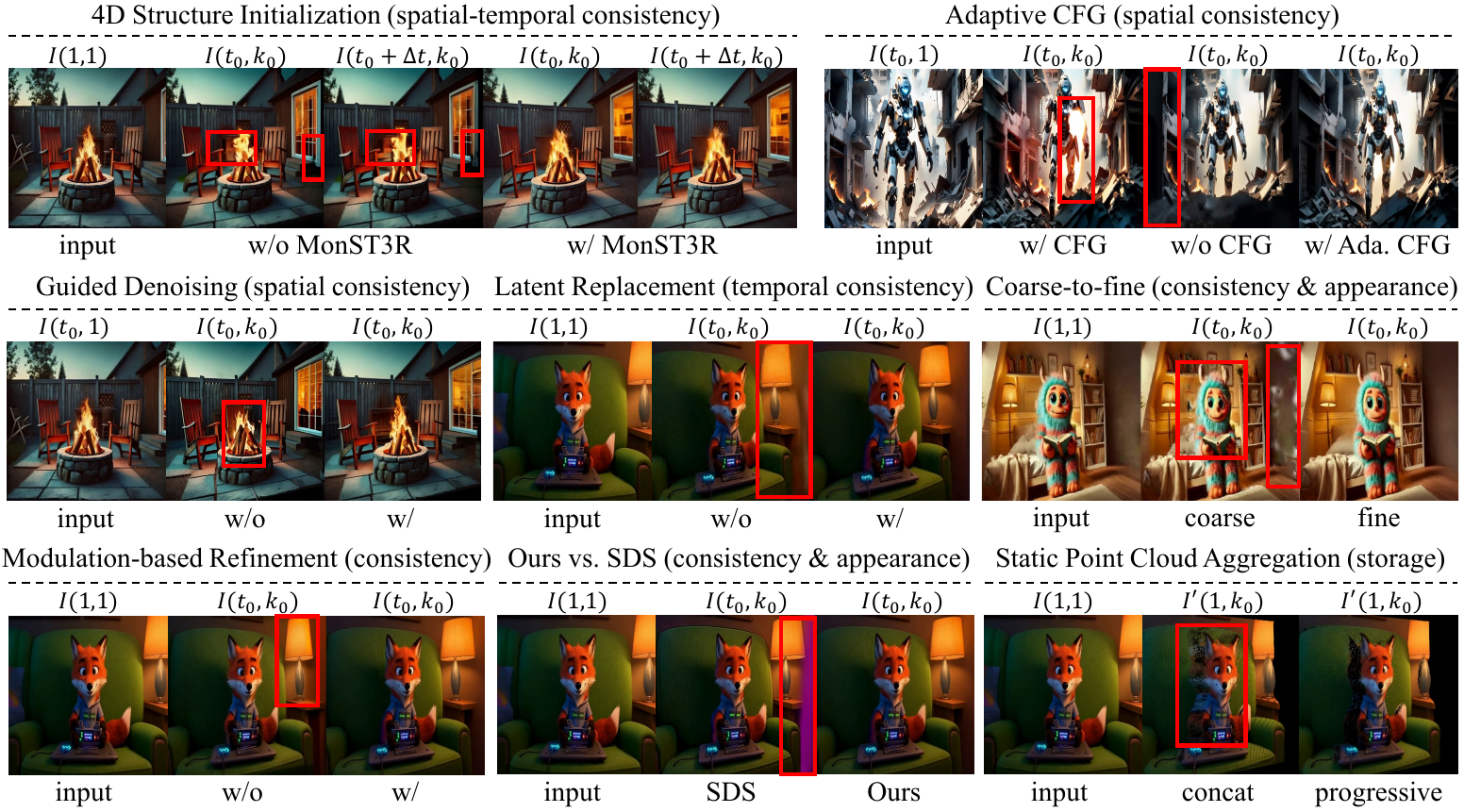}
    \vspace{-5pt}
    \caption{\textbf{Qualitative Comparison of Ablation Studies.}}
    \label{fig:ablation}
    \vspace{-10pt}
\end{figure*}

\begin{table}[!t]
    \small
  \centering
  \setlength{\tabcolsep}{3pt}
  \begin{tabular}{c|cccc}
    \toprule
     Method & Text Align & Consistency & Dynamic & Aesthetic \\ \midrule
     4Real~\cite{4real}   
     & 26.1\%    & 95.7\%    & 32.3\%  & 50.9\%  \\
     Ours
     & 26.1\%    & 96.0\%    & 47.4\%  & 64.7\%  \\
     \midrule
     4Dfy~\cite{4dfy}   
     & 25.7\%    & 91.6\%    & 53.3\% & 54.5\%  \\
     Ours
     & 26.0\%    & 96.9\%    & 54.1\% & 61.9\%  \\
     \midrule
     D-in-4D~\cite{dream-in-4d}   
     & 25.0\%    & 91.0\%    & 53.5\% & 55.1\% \\
     Ours
     & 25.9\%     & 95.2\%    & 53.2\% & 65.3\% \\
     \bottomrule
    \end{tabular}
    \vspace{-5pt}
    \caption{
    \textbf{Text-to-4D comparisons on VBench~\cite{vbench}.} We report the text alignment, consistency, dynamics, and aesthetics of the generated 4D videos. D-in-4D denotes Dream-in-4D~\cite{dream-in-4d}.
    }
    \vspace{-10pt}
  \label{tab:text-benchmark}
\end{table}

\begin{table}[!t]
    \small
  \centering
  \setlength{\tabcolsep}{7pt}
  \begin{tabular}{c|ccc}
    \toprule
     Method & Consistency & Dynamic & Aesthetic \\ \midrule
     Animate124~\cite{animate124}   
     & 90.7\%    & 45.4\%    & 42.3\%  \\
     Ours
     & 96.9\%    & 40.1\%    & 60.5\%  \\
     \midrule
     DimensionX~\cite{dimensionX}   
     & 97.2\%      & 21.9\%    & 56.0\%  \\
     Ours
     & 95.5\%      & 22.1\%    & 57.3\%  \\
     \midrule
     GenXD~\cite{genXD}   
     & 89.8\%    & 98.3\%   & 38.0\% \\
     Ours
     & 96.8\%    & 100.0\%  & 57.9\% \\
     \bottomrule
    \end{tabular}
    \vspace{-5pt}
    \caption{
    \textbf{Image-to-4D comparisons on VBench~\cite{vbench}.} We report the text alignment, consistency, dynamics, and aesthetics of the generated 4D videos.
    }
    \vspace{-10pt}
  \label{tab:image-benchmark-vbench}
\end{table}

\begin{figure*}[!t]
  \centering
    \includegraphics[width=0.99\linewidth]{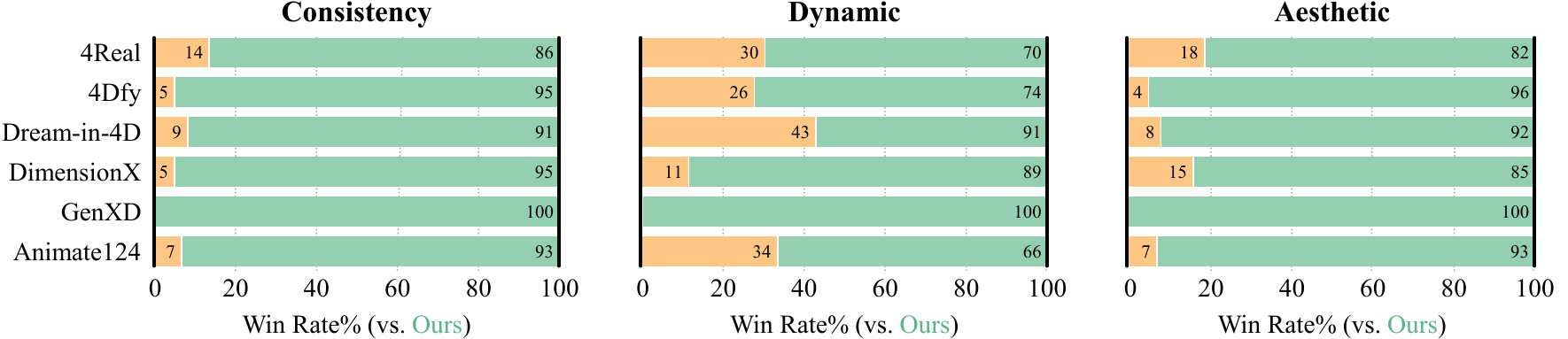}
    \vspace{-5pt}
    \caption{\textbf{Comparison of different methods based on the user study.}
    }
    \label{fig:user study}
    \vspace{-10pt}
\end{figure*}

\noindent \textbf{Datasets and Metrics.}
The data used for evaluation, including single images and texts, are sourced from the project pages of the comparison methods.
To evaluate the quality of multi-view videos rendered from 4D representations, we report common VBench~\cite{vbench} metrics: Consistency (average for subject/background), Dynamic Degree, Aesthetic Score, and Text Alignment (only for text-to-4D). 
Since there is no well-established benchmark in 4D generation field, we conduct a user study with $78$ evaluators to enhance reliability. 
Details on these metrics and the user study are provided in~\Cref{supp:details-of-user-study}.

\noindent \textbf{Implementation Details.}
We use~\cite{4dgs} as our 4D representation. In the coarse stage, we first train for $9$k iterations, followed by $1$k iterations in the fine stage. We conduct all experiments on a single NVIDIA A100~(40GB) GPU. More details on the network, hyperparameter settings, and runtime are provided in~\Cref{supp:more-implementation-details}.

\subsection{Text-to-4D Comparisons} 
The qualitative comparisons are presented in \Cref{fig:comparison-text}, while the quantitative results on VBench~\cite{vbench} and the user study are shown in~\Cref{tab:text-benchmark} and~\Cref{fig:user study}, respectively.
On VBench~(\Cref{tab:text-benchmark}), our method is comparable to or outperforms 4Real~\cite{4real} across all three dimensions, with notable improvements in Aesthetics and Dynamics. Compared to object-level methods, our approach excels in Consistency and Aesthetics while performing comparably in Dynamics, slightly lower than Dream-in-4D~\cite{dream-in-4d}. 
This is mainly because object-level generation can more easily handle large viewpoint shifts, as it focuses only on simple single objects without complex textures or backgrounds.
In contrast, our \methodname can generate more dynamic scenes with complex textures and backgrounds, as illustrated in~\Cref{fig:comparison-text}.
Furthermore, the user study~(\Cref{fig:user study}) provides strong evidence of \methodname's superiority. Evaluators consistently found our generated results to be more advantageous across all four dimensions compared to other methods. This demonstrates that \methodname not only produces more aesthetically pleasing videos but also generates results with greater diversity, coherence, and realism. Overall, these findings highlight the effectiveness and robustness of our proposed \methodname in scene generation.

\subsection{Image-to-4D Comparisons} 
\Cref{tab:image-benchmark-vbench} and \Cref{fig:user study} present the quantitative comparisons on VBench~\cite{vbench} and user studies, while qualitative results are shown in \Cref{fig:comparison-img}.
Compared to GenXD~\cite{genXD}, \methodname achieves more realistic and consistent free-viewpoint video reconstruction from a single image. This is more evident in the quantitative VBench~\cite{vbench} and is further corroborated by the qualitative results in~\Cref{fig:comparison-img}. 
Compared to the object-centric Animate124~\cite{animate124}, our proposed scene-level \methodname not only incorporates the environment but also exhibits fewer artifacts and better temporal consistency. This highlights \methodname’s ability to generate high-quality, coherent scenes with complex textures and backgrounds, a common challenge for object-centric methods.
Furthermore, \methodname achieves results comparable to DimensionX~\cite{dimensionX} on the VBench~\cite{vbench} benchmark while outperforming it on user preferences. 
Evaluators in the user study consistently favored our \methodname for its diversity, consistency, and realism. This further underscores the effectiveness of our method in generating high-quality, free-viewpoint videos from single images without the need for tuning. 
More results can be found on our project page.

\begin{table}[!t]
    \small
  \centering
  \setlength{\tabcolsep}{5pt}
  \begin{tabular}{l|ccc}
    \toprule
     Method & Consistency & Dynamic & Aesthetic \\ \midrule
     wo / w MonST3R
     & 14\% / 86\% & 30\% / 70\% & 9\% / 91\% \\
     wo / w Ada. CFG
     & 14\% / 86\% & 36\% / 64\% & 25\% / 75\% \\
     wo / w PGD$^*$
     & 14\% / 86\% & 11\% / 89\% & 13\% / 87\% \\
     wo / w RLR$^\dag$
     & 24\% / 76\% & 31\% / 69\% & 17\% / 83\% \\
     wo / w Fine Stage
     & 4\% / 96\% & 21\% / 79\% & 6\% / 94\% \\
     wo / w MBR$^\ddag$
     & 5\% / 95\% & 14\% / 86\% & 6\% / 94\% \\
     SDS vs. Ours
     & 8\% / 92\% & 10\% / 90\% & 9\% / 91\% \\
     \bottomrule
    \end{tabular}
    \vspace{-5pt}
    \caption{\textbf{User study results on ablations.} PGD$^*$, RLR$^\dag$, and MBR$^\ddag$ refer to point cloud guided denoising, reference latent replacement, and modulation-based refinement, respectively.
    }
    \vspace{-10pt}
  \label{tab:user-study}
\end{table}

\subsection{Ablations and Analysis} 
\label{subsec:ablations}
We analyze our pipeline by systematically removing individual components and evaluating their impact.
\Cref{fig:ablation} presents the quantitative results, while~\Cref{tab:user-study} includes the corresponding user studies. 

\noindent\textbf{MonST3R provides effective 4D structure initialization.}
The integration of MonST3R~\cite{monst3r} for 4D structure construction is crucial for preserving geometric and spatial consistency, outperforming~\cite{dust3r} used by~\cite{viewcrafter}.

\noindent\textbf{Adaptive CFG enhances view consistency.} Standard CFG would introduce noticeable color shifts between views, sometimes leading to oversaturation, while disabling it weakens completion in missing regions. Our proposed Adaptive CFG achieves a well-balanced trade-off, enhancing consistency across views.

\noindent\textbf{Point Cloud Guided Denoising mitigates unexpected motion.} This technique stabilizes dynamic subjects, such as fluids, ensuring consistency across different views. Without this module, undesired fluid dynamics may occur.

\noindent\textbf{Reference Latent Replacement is crucial for temporal consistency.} 
Without it, generated results in occluded and missing regions exhibit significant variations across different time steps for the same viewpoint, leading to temporal inconsistencies and blurring.

\noindent\textbf{Refinement with multi-view video significantly improves consistency and appearance.} Without refinement, the coarse-stage results exhibit noticeable artifacts and blurring, while refinement greatly enhances overall quality.

\noindent\textbf{Modulation-based Refinement aggregates generated information while preserving consistency.} Direct supervision with generated multi-view videos can introduce temporal inconsistencies. Additionally, pixel-level SDS leads to unstable training and oversaturation in missing regions.

\noindent\textbf{Progressive static point aggregation preserves background integrity while minimizing storage.} Directly concatenating background point clouds from all frames leads to excessive data size and introduces ghosting artifacts.

\section{Conclusion}
\label{sec:conclusion}
We introduce \methodname, the first tuning-free approach for generating consistent 4D scenes from a single image. Our approach begins with a 4D geometric structure construction module to initialize multi-view videos, followed by a point-based generative model. To ensure spatial-temporal consistency, we incorporate adaptive classifier-free guidance and a point cloud guided denoising strategy for spatial coherence, along with reference latent replacement for temporal consistency. Finally, we apply an effective training strategy with a modulation-based refinement to lift the generated multi-view video into a consistent 4D representation.

{
    \small
    \bibliographystyle{ieeenat_fullname}
    \bibliography{main}

\begin{thebibliography}{81}
\providecommand{\natexlab}[1]{#1}
\providecommand{\url}[1]{\texttt{#1}}
\expandafter\ifx\csname urlstyle\endcsname\relax
  \providecommand{\doi}[1]{doi: #1}\else
  \providecommand{\doi}{doi: \begingroup \urlstyle{rm}\Url}\fi

\bibitem[Achlioptas et~al.(2018)Achlioptas, Diamanti, Mitliagkas, and Guibas]{DBLP:conf/iclr/AchlioptasDMG18}
Panos Achlioptas, Olga Diamanti, Ioannis Mitliagkas, and Leonidas~J. Guibas.
\newblock Learning representations and generative models for 3d point clouds.
\newblock In \emph{6th International Conference on Learning Representations, {ICLR} 2018, Vancouver, BC, Canada, April 30 - May 3, 2018, Workshop Track Proceedings}. OpenReview.net, 2018.

\bibitem[Babaeizadeh et~al.(2018)Babaeizadeh, Finn, Erhan, Campbell, and Levine]{DBLP:conf/iclr/BabaeizadehFECL18}
Mohammad Babaeizadeh, Chelsea Finn, Dumitru Erhan, Roy~H. Campbell, and Sergey Levine.
\newblock Stochastic variational video prediction.
\newblock In \emph{6th International Conference on Learning Representations, {ICLR} 2018, Vancouver, BC, Canada, April 30 - May 3, 2018, Conference Track Proceedings}. OpenReview.net, 2018.

\bibitem[Babaeizadeh et~al.(2021)Babaeizadeh, Saffar, Nair, Levine, Finn, and Erhan]{babaeizadeh2021fitvid}
Mohammad Babaeizadeh, Mohammad~Taghi Saffar, Suraj Nair, Sergey Levine, Chelsea Finn, and Dumitru Erhan.
\newblock Fitvid: Overfitting in pixel-level video prediction.
\newblock \emph{arXiv preprint arXiv:2106.13195}, 2021.

\bibitem[Bahmani et~al.(2024)Bahmani, Skorokhodov, Rong, Wetzstein, Guibas, Wonka, Tulyakov, Park, Tagliasacchi, and Lindell]{4dfy}
Sherwin Bahmani, Ivan Skorokhodov, Victor Rong, Gordon Wetzstein, Leonidas~J. Guibas, Peter Wonka, Sergey Tulyakov, Jeong~Joon Park, Andrea Tagliasacchi, and David~B. Lindell.
\newblock 4d-fy: Text-to-4d generation using hybrid score distillation sampling.
\newblock In \emph{{IEEE/CVF} Conference on Computer Vision and Pattern Recognition, {CVPR} 2024, Seattle, WA, USA, June 16-22, 2024}, pages 7996--8006. {IEEE}, 2024.

\bibitem[Bai et~al.(2024)Bai, Xia, Wang, Yuan, Fu, Liu, Hu, Wan, and Zhang]{SynCamMaster}
Jianhong Bai, Menghan Xia, Xintao Wang, Ziyang Yuan, Xiao Fu, Zuozhu Liu, Haoji Hu, Pengfei Wan, and Di Zhang.
\newblock Syncammaster: Synchronizing multi-camera video generation from diverse viewpoints.
\newblock \emph{CoRR}, abs/2412.07760, 2024.

\bibitem[Blattmann et~al.(2021)Blattmann, Milbich, Dorkenwald, and Ommer]{blattmann2021understanding}
Andreas Blattmann, Timo Milbich, Michael Dorkenwald, and Bjorn Ommer.
\newblock Understanding object dynamics for interactive image-to-video synthesis.
\newblock In \emph{Proceedings of the IEEE/CVF Conference on Computer Vision and Pattern Recognition}, pages 5171--5181, 2021.

\bibitem[Blattmann et~al.(2023)Blattmann, Rombach, Ling, Dockhorn, Kim, Fidler, and Kreis]{DBLP:conf/cvpr/BlattmannRLD0FK23}
Andreas Blattmann, Robin Rombach, Huan Ling, Tim Dockhorn, Seung~Wook Kim, Sanja Fidler, and Karsten Kreis.
\newblock Align your latents: High-resolution video synthesis with latent diffusion models.
\newblock In \emph{{IEEE/CVF} Conference on Computer Vision and Pattern Recognition, {CVPR} 2023, Vancouver, BC, Canada, June 17-24, 2023}, pages 22563--22575. {IEEE}, 2023.

\bibitem[Caron et~al.(2021)Caron, Touvron, Misra, J{\'{e}}gou, Mairal, Bojanowski, and Joulin]{dino}
Mathilde Caron, Hugo Touvron, Ishan Misra, Herv{\'{e}} J{\'{e}}gou, Julien Mairal, Piotr Bojanowski, and Armand Joulin.
\newblock Emerging properties in self-supervised vision transformers.
\newblock In \emph{2021 {IEEE/CVF} International Conference on Computer Vision, {ICCV} 2021, Montreal, QC, Canada, October 10-17, 2021}, pages 9630--9640. {IEEE}, 2021.

\bibitem[Chen et~al.(2024{\natexlab{a}})Chen, Zhang, Cun, Xia, Wang, Weng, and Shan]{videocrafter2}
Haoxin Chen, Yong Zhang, Xiaodong Cun, Menghan Xia, Xintao Wang, Chao Weng, and Ying Shan.
\newblock Videocrafter2: Overcoming data limitations for high-quality video diffusion models.
\newblock In \emph{{IEEE/CVF} Conference on Computer Vision and Pattern Recognition, {CVPR} 2024, Seattle, WA, USA, June 16-22, 2024}, pages 7310--7320. {IEEE}, 2024{\natexlab{a}}.

\bibitem[Chen et~al.(2023)Chen, Wang, and Liu]{DBLP:journals/pami/ChenWL23}
Zhaoxi Chen, Guangcong Wang, and Ziwei Liu.
\newblock Scenedreamer: Unbounded 3d scene generation from 2d image collections.
\newblock \emph{{IEEE} Trans. Pattern Anal. Mach. Intell.}, 45\penalty0 (12):\penalty0 15562--15576, 2023.

\bibitem[Chen et~al.(2024{\natexlab{b}})Chen, Tang, Dong, Cao, Hong, Lan, Wang, Xie, Wu, Saito, Pan, Lin, and Liu]{DBLP:journals/corr/abs-2409-12957}
Zhaoxi Chen, Jiaxiang Tang, Yuhao Dong, Ziang Cao, Fangzhou Hong, Yushi Lan, Tengfei Wang, Haozhe Xie, Tong Wu, Shunsuke Saito, Liang Pan, Dahua Lin, and Ziwei Liu.
\newblock 3dtopia-xl: Scaling high-quality 3d asset generation via primitive diffusion.
\newblock \emph{CoRR}, abs/2409.12957, 2024{\natexlab{b}}.

\bibitem[Chou et~al.(2023)Chou, Bahat, and Heide]{DBLP:conf/iccv/ChouBH23}
Gene Chou, Yuval Bahat, and Felix Heide.
\newblock Diffusion-sdf: Conditional generative modeling of signed distance functions.
\newblock In \emph{{IEEE/CVF} International Conference on Computer Vision, {ICCV} 2023, Paris, France, October 1-6, 2023}, pages 2262--2272. {IEEE}, 2023.

\bibitem[Clark et~al.(2019)Clark, Donahue, and Simonyan]{clark2019adversarial}
Aidan Clark, Jeff Donahue, and Karen Simonyan.
\newblock Adversarial video generation on complex datasets.
\newblock \emph{arXiv preprint arXiv:1907.06571}, 2019.

\bibitem[Denton and Fergus(2018)]{DBLP:conf/icml/DentonF18}
Emily Denton and Rob Fergus.
\newblock Stochastic video generation with a learned prior.
\newblock In \emph{Proceedings of the 35th International Conference on Machine Learning, {ICML} 2018, Stockholmsm{\"{a}}ssan, Stockholm, Sweden, July 10-15, 2018}, pages 1182--1191. {PMLR}, 2018.

\bibitem[Gao et~al.(2022)Gao, Shen, Wang, Chen, Yin, Li, Litany, Gojcic, and Fidler]{get3d}
Jun Gao, Tianchang Shen, Zian Wang, Wenzheng Chen, Kangxue Yin, Daiqing Li, Or Litany, Zan Gojcic, and Sanja Fidler.
\newblock {GET3D:} {A} generative model of high quality 3d textured shapes learned from images.
\newblock In \emph{Advances in Neural Information Processing Systems 35: Annual Conference on Neural Information Processing Systems 2022, NeurIPS 2022, New Orleans, LA, USA, November 28 - December 9, 2022}, 2022.

\bibitem[Ge et~al.(2023)Ge, Nah, Liu, Poon, Tao, Catanzaro, Jacobs, Huang, Liu, and Balaji]{DBLP:conf/iccv/GeNLPTC0H0B23}
Songwei Ge, Seungjun Nah, Guilin Liu, Tyler Poon, Andrew Tao, Bryan Catanzaro, David Jacobs, Jia{-}Bin Huang, Ming{-}Yu Liu, and Yogesh Balaji.
\newblock Preserve your own correlation: {A} noise prior for video diffusion models.
\newblock In \emph{{IEEE/CVF} International Conference on Computer Vision, {ICCV} 2023, Paris, France, October 1-6, 2023}, pages 22873--22884. {IEEE}, 2023.

\bibitem[Guo et~al.(2024)Guo, Yang, Rao, Liang, Wang, Qiao, Agrawala, Lin, and Dai]{DBLP:conf/iclr/0002YRL00AL024}
Yuwei Guo, Ceyuan Yang, Anyi Rao, Zhengyang Liang, Yaohui Wang, Yu Qiao, Maneesh Agrawala, Dahua Lin, and Bo Dai.
\newblock Animatediff: Animate your personalized text-to-image diffusion models without specific tuning.
\newblock In \emph{The Twelfth International Conference on Learning Representations, {ICLR} 2024, Vienna, Austria, May 7-11, 2024}. OpenReview.net, 2024.

\bibitem[Harvey et~al.(2022)Harvey, Naderiparizi, Masrani, Weilbach, and Wood]{harvey2022flexible}
William Harvey, Saeid Naderiparizi, Vaden Masrani, Christian Weilbach, and Frank Wood.
\newblock Flexible diffusion modeling of long videos.
\newblock \emph{Advances in Neural Information Processing Systems}, 35:\penalty0 27953--27965, 2022.

\bibitem[He et~al.(2022)He, Yang, Zhang, Shan, and Chen]{DBLP:journals/corr/abs-2211-13221}
Yingqing He, Tianyu Yang, Yong Zhang, Ying Shan, and Qifeng Chen.
\newblock Latent video diffusion models for high-fidelity video generation with arbitrary lengths.
\newblock \emph{CoRR}, abs/2211.13221, 2022.

\bibitem[Ho and Salimans(2022)]{cfg}
Jonathan Ho and Tim Salimans.
\newblock Classifier-free diffusion guidance.
\newblock \emph{arXiv preprint arXiv:2207.12598}, 2022.

\bibitem[Ho et~al.(2022{\natexlab{a}})Ho, Chan, Saharia, Whang, Gao, Gritsenko, Kingma, Poole, Norouzi, Fleet, and Salimans]{DBLP:journals/corr/abs-2210-02303}
Jonathan Ho, William Chan, Chitwan Saharia, Jay Whang, Ruiqi Gao, Alexey~A. Gritsenko, Diederik~P. Kingma, Ben Poole, Mohammad Norouzi, David~J. Fleet, and Tim Salimans.
\newblock Imagen video: High definition video generation with diffusion models.
\newblock \emph{CoRR}, abs/2210.02303, 2022{\natexlab{a}}.

\bibitem[Ho et~al.(2022{\natexlab{b}})Ho, Salimans, Gritsenko, Chan, Norouzi, and Fleet]{VDM}
Jonathan Ho, Tim Salimans, Alexey Gritsenko, William Chan, Mohammad Norouzi, and David~J Fleet.
\newblock Video diffusion models.
\newblock \emph{Advances in Neural Information Processing Systems}, 35:\penalty0 8633--8646, 2022{\natexlab{b}}.

\bibitem[Hong et~al.(2023)Hong, Ding, Zheng, Liu, and Tang]{cogvideo}
Wenyi Hong, Ming Ding, Wendi Zheng, Xinghan Liu, and Jie Tang.
\newblock Cogvideo: Large-scale pretraining for text-to-video generation via transformers.
\newblock In \emph{The Eleventh International Conference on Learning Representations, {ICLR} 2023, Kigali, Rwanda, May 1-5, 2023}. OpenReview.net, 2023.

\bibitem[Hu et~al.(2022)Hu, Luo, and Chen]{DBLP:conf/cvpr/HuLC22}
Yaosi Hu, Chong Luo, and Zhenzhong Chen.
\newblock Make it move: Controllable image-to-video generation with text descriptions.
\newblock In \emph{{IEEE/CVF} Conference on Computer Vision and Pattern Recognition, {CVPR} 2022, New Orleans, LA, USA, June 18-24, 2022}, pages 18198--18207. {IEEE}, 2022.

\bibitem[Huang et~al.(2024{\natexlab{a}})Huang, Yu, Chen, Geiger, and Gao]{2dgs}
Binbin Huang, Zehao Yu, Anpei Chen, Andreas Geiger, and Shenghua Gao.
\newblock 2d gaussian splatting for geometrically accurate radiance fields.
\newblock In \emph{{ACM} {SIGGRAPH} 2024 Conference Papers, {SIGGRAPH} 2024, Denver, CO, USA, 27 July 2024- 1 August 2024}, page~32. {ACM}, 2024{\natexlab{a}}.

\bibitem[Huang et~al.(2024{\natexlab{b}})Huang, He, Yu, Zhang, Si, Jiang, Zhang, Wu, Jin, Chanpaisit, Wang, Chen, Wang, Lin, Qiao, and Liu]{vbench}
Ziqi Huang, Yinan He, Jiashuo Yu, Fan Zhang, Chenyang Si, Yuming Jiang, Yuanhan Zhang, Tianxing Wu, Qingyang Jin, Nattapol Chanpaisit, Yaohui Wang, Xinyuan Chen, Limin Wang, Dahua Lin, Yu Qiao, and Ziwei Liu.
\newblock Vbench: Comprehensive benchmark suite for video generative models.
\newblock In \emph{{IEEE/CVF} Conference on Computer Vision and Pattern Recognition, {CVPR} 2024, Seattle, WA, USA, June 16-22, 2024}, pages 21807--21818. {IEEE}, 2024{\natexlab{b}}.

\bibitem[Kalchbrenner et~al.(2017)Kalchbrenner, van~den Oord, Simonyan, Danihelka, Vinyals, Graves, and Kavukcuoglu]{DBLP:conf/icml/KalchbrennerOSD17}
Nal Kalchbrenner, A{\"{a}}ron van~den Oord, Karen Simonyan, Ivo Danihelka, Oriol Vinyals, Alex Graves, and Koray Kavukcuoglu.
\newblock Video pixel networks.
\newblock In \emph{Proceedings of the 34th International Conference on Machine Learning, {ICML} 2017, Sydney, NSW, Australia, 6-11 August 2017}, pages 1771--1779. {PMLR}, 2017.

\bibitem[Kerbl et~al.(2023)Kerbl, Kopanas, Leimk{\"{u}}hler, and Drettakis]{3dgs}
Bernhard Kerbl, Georgios Kopanas, Thomas Leimk{\"{u}}hler, and George Drettakis.
\newblock 3d gaussian splatting for real-time radiance field rendering.
\newblock \emph{{ACM} Trans. Graph.}, 42\penalty0 (4):\penalty0 139:1--139:14, 2023.

\bibitem[Kuang et~al.(2024)Kuang, Cai, He, Xu, Li, Guibas, and Wetzstein]{Collaborative-Video-Diffusion}
Zhengfei Kuang, Shengqu Cai, Hao He, Yinghao Xu, Hongsheng Li, Leonidas~J. Guibas, and Gordon Wetzstein.
\newblock Collaborative video diffusion: Consistent multi-video generation with camera control.
\newblock In \emph{Advances in Neural Information Processing Systems 38: Annual Conference on Neural Information Processing Systems 2024, NeurIPS 2024, Vancouver, BC, Canada, December 10 - 15, 2024}, 2024.

\bibitem[Kumar et~al.(2020)Kumar, Babaeizadeh, Erhan, Finn, Levine, Dinh, and Kingma]{DBLP:conf/iclr/KumarBEFLDK20}
Manoj Kumar, Mohammad Babaeizadeh, Dumitru Erhan, Chelsea Finn, Sergey Levine, Laurent Dinh, and Durk Kingma.
\newblock Videoflow: {A} conditional flow-based model for stochastic video generation.
\newblock In \emph{8th International Conference on Learning Representations, {ICLR} 2020, Addis Ababa, Ethiopia, April 26-30, 2020}. OpenReview.net, 2020.

\bibitem[LAION-AI(2022)]{LAION}
LAION-AI.
\newblock aesthetic-predictor, 2022.

\bibitem[Li et~al.(2018)Li, Fang, Yang, Wang, Lu, and Yang]{DBLP:conf/eccv/LiFYWLY18}
Yijun Li, Chen Fang, Jimei Yang, Zhaowen Wang, Xin Lu, and Ming{-}Hsuan Yang.
\newblock Flow-grounded spatial-temporal video prediction from still images.
\newblock In \emph{Computer Vision - {ECCV} 2018 - 15th European Conference, Munich, Germany, September 8-14, 2018, Proceedings, Part {IX}}, pages 609--625. Springer, 2018.

\bibitem[Lin et~al.(2023)Lin, Lee, Menapace, Chai, Siarohin, Yang, and Tulyakov]{DBLP:conf/iccv/LinLMCS0T23}
Chieh~Hubert Lin, Hsin{-}Ying Lee, Willi Menapace, Menglei Chai, Aliaksandr Siarohin, Ming{-}Hsuan Yang, and Sergey Tulyakov.
\newblock Infinicity: Infinite-scale city synthesis.
\newblock In \emph{{IEEE/CVF} International Conference on Computer Vision, {ICCV} 2023, Paris, France, October 1-6, 2023}, pages 22751--22761. {IEEE}, 2023.

\bibitem[Lin et~al.(2024)Lin, Liu, Li, and Yang]{stdalign}
Shanchuan Lin, Bingchen Liu, Jiashi Li, and Xiao Yang.
\newblock Common diffusion noise schedules and sample steps are flawed.
\newblock In \emph{WACV}, pages 5404--5411, 2024.

\bibitem[Luiten et~al.(2024)Luiten, Kopanas, Leibe, and Ramanan]{dy3dgs}
Jonathon Luiten, Georgios Kopanas, Bastian Leibe, and Deva Ramanan.
\newblock Dynamic 3d gaussians: Tracking by persistent dynamic view synthesis.
\newblock In \emph{International Conference on 3D Vision, 3DV 2024, Davos, Switzerland, March 18-21, 2024}, pages 800--809. {IEEE}, 2024.

\bibitem[Luo et~al.(2023)Luo, Chen, Zhang, Huang, Wang, Shen, Zhao, Zhou, and Tan]{videofusion}
Zhengxiong Luo, Dayou Chen, Yingya Zhang, Yan Huang, Liang Wang, Yujun Shen, Deli Zhao, Jingren Zhou, and Tieniu Tan.
\newblock Videofusion: Decomposed diffusion models for high-quality video generation.
\newblock \emph{CoRR}, abs/2303.08320, 2023.

\bibitem[Mathieu et~al.(2016)Mathieu, Couprie, and LeCun]{DBLP:journals/corr/MathieuCL15}
Micha{\"{e}}l Mathieu, Camille Couprie, and Yann LeCun.
\newblock Deep multi-scale video prediction beyond mean square error.
\newblock In \emph{4th International Conference on Learning Representations, {ICLR} 2016, San Juan, Puerto Rico, May 2-4, 2016, Conference Track Proceedings}, 2016.

\bibitem[Menapace et~al.(2024)Menapace, Siarohin, Skorokhodov, Deyneka, Chen, Kag, Fang, Stoliar, Ricci, Ren, et~al.]{snap}
Willi Menapace, Aliaksandr Siarohin, Ivan Skorokhodov, Ekaterina Deyneka, Tsai-Shien Chen, Anil Kag, Yuwei Fang, Aleksei Stoliar, Elisa Ricci, Jian Ren, et~al.
\newblock Snap video: Scaled spatiotemporal transformers for text-to-video synthesis.
\newblock In \emph{Proceedings of the IEEE/CVF Conference on Computer Vision and Pattern Recognition}, pages 7038--7048, 2024.

\bibitem[Mildenhall et~al.(2022)Mildenhall, Srinivasan, Tancik, Barron, Ramamoorthi, and Ng]{nerf}
Ben Mildenhall, Pratul~P. Srinivasan, Matthew Tancik, Jonathan~T. Barron, Ravi Ramamoorthi, and Ren Ng.
\newblock Nerf: representing scenes as neural radiance fields for view synthesis.
\newblock \emph{Commun. {ACM}}, 65\penalty0 (1):\penalty0 99--106, 2022.

\bibitem[Mokady et~al.(2023)Mokady, Hertz, Aberman, Pritch, and Cohen-Or]{mokady2023null}
Ron Mokady, Amir Hertz, Kfir Aberman, Yael Pritch, and Daniel Cohen-Or.
\newblock Null-text inversion for editing real images using guided diffusion models.
\newblock In \emph{Proceedings of the IEEE/CVF conference on computer vision and pattern recognition}, pages 6038--6047, 2023.

\bibitem[Pan et~al.(2019)Pan, Wang, Jia, Shao, Sheng, Yan, and Wang]{pan2019video}
Junting Pan, Chengyu Wang, Xu Jia, Jing Shao, Lu Sheng, Junjie Yan, and Xiaogang Wang.
\newblock Video generation from single semantic label map.
\newblock In \emph{Proceedings of the IEEE/CVF Conference on Computer Vision and Pattern Recognition}, pages 3733--3742, 2019.

\bibitem[Pan et~al.(2024)Pan, Yang, Zhu, and Zhang]{pan2024efficient4d}
Zijie Pan, Zeyu Yang, Xiatian Zhu, and Li Zhang.
\newblock Efficient4d: Fast dynamic 3d object generation from a single-view video.
\newblock \emph{arXiv preprint arXiv:2401.08742}, 2024.

\bibitem[Park et~al.(2019)Park, Florence, Straub, Newcombe, and Lovegrove]{DBLP:conf/cvpr/ParkFSNL19}
Jeong~Joon Park, Peter~R. Florence, Julian Straub, Richard~A. Newcombe, and Steven Lovegrove.
\newblock Deepsdf: Learning continuous signed distance functions for shape representation.
\newblock In \emph{{IEEE} Conference on Computer Vision and Pattern Recognition, {CVPR} 2019, Long Beach, CA, USA, June 16-20, 2019}, pages 165--174. Computer Vision Foundation / {IEEE}, 2019.

\bibitem[Poole et~al.(2022)Poole, Jain, Barron, and Mildenhall]{dreamfusion}
Ben Poole, Ajay Jain, Jonathan~T Barron, and Ben Mildenhall.
\newblock Dreamfusion: Text-to-3d using 2d diffusion.
\newblock \emph{arXiv preprint arXiv:2209.14988}, 2022.

\bibitem[Pumarola et~al.(2021)Pumarola, Corona, Pons{-}Moll, and Moreno{-}Noguer]{d-nerf}
Albert Pumarola, Enric Corona, Gerard Pons{-}Moll, and Francesc Moreno{-}Noguer.
\newblock D-nerf: Neural radiance fields for dynamic scenes.
\newblock In \emph{{IEEE} Conference on Computer Vision and Pattern Recognition, {CVPR} 2021, virtual, June 19-25, 2021}, pages 10318--10327. Computer Vision Foundation / {IEEE}, 2021.

\bibitem[Radford et~al.(2021)Radford, Kim, Hallacy, Ramesh, Goh, Agarwal, Sastry, Askell, Mishkin, Clark, Krueger, and Sutskever]{clip}
Alec Radford, Jong~Wook Kim, Chris Hallacy, Aditya Ramesh, Gabriel Goh, Sandhini Agarwal, Girish Sastry, Amanda Askell, Pamela Mishkin, Jack Clark, Gretchen Krueger, and Ilya Sutskever.
\newblock Learning transferable visual models from natural language supervision.
\newblock In \emph{Proceedings of the 38th International Conference on Machine Learning, {ICML} 2021, 18-24 July 2021, Virtual Event}, pages 8748--8763. {PMLR}, 2021.

\bibitem[Rahamim et~al.(2024)Rahamim, Malca, Samuel, and Chechik]{DBLP:journals/corr/abs-2412-20422}
Ohad Rahamim, Ori Malca, Dvir Samuel, and Gal Chechik.
\newblock Bringing objects to life: 4d generation from 3d objects.
\newblock \emph{CoRR}, abs/2412.20422, 2024.

\bibitem[Ranzato et~al.(2014)Ranzato, Szlam, Bruna, Mathieu, Collobert, and Chopra]{DBLP:journals/corr/RanzatoSBMCC14}
Marc'Aurelio Ranzato, Arthur Szlam, Joan Bruna, Micha{\"{e}}l Mathieu, Ronan Collobert, and Sumit Chopra.
\newblock Video (language) modeling: a baseline for generative models of natural videos.
\newblock \emph{CoRR}, abs/1412.6604, 2014.

\bibitem[Ren et~al.(2023)Ren, Pan, Tang, Zhang, Cao, Zeng, and Liu]{dreamgaussian4d}
Jiawei Ren, Liang Pan, Jiaxiang Tang, Chi Zhang, Ang Cao, Gang Zeng, and Ziwei Liu.
\newblock Dreamgaussian4d: Generative 4d gaussian splatting.
\newblock \emph{arXiv preprint arXiv:2312.17142}, 2023.

\bibitem[Ren et~al.(2025)Ren, Xie, Mirzaei, Kreis, Liu, Torralba, Fidler, Kim, Ling, et~al.]{l4gm}
Jiawei Ren, Cheng Xie, Ashkan Mirzaei, Karsten Kreis, Ziwei Liu, Antonio Torralba, Sanja Fidler, Seung~Wook Kim, Huan Ling, et~al.
\newblock L4gm: Large 4d gaussian reconstruction model.
\newblock \emph{Advances in Neural Information Processing Systems}, 37:\penalty0 56828--56858, 2025.

\bibitem[Rombach et~al.(2022)Rombach, Blattmann, Lorenz, Esser, and Ommer]{ldm}
Robin Rombach, Andreas Blattmann, Dominik Lorenz, Patrick Esser, and Bj{\"o}rn Ommer.
\newblock High-resolution image synthesis with latent diffusion models.
\newblock In \emph{Proceedings of the IEEE/CVF conference on computer vision and pattern recognition}, pages 10684--10695, 2022.

\bibitem[Schonberger and Frahm(2016)]{colmap}
Johannes~L Schonberger and Jan-Michael Frahm.
\newblock Structure-from-motion revisited.
\newblock In \emph{Proceedings of the IEEE conference on computer vision and pattern recognition}, pages 4104--4113, 2016.

\bibitem[Singer et~al.(2023{\natexlab{a}})Singer, Polyak, Hayes, Yin, An, Zhang, Hu, Yang, Ashual, Gafni, Parikh, Gupta, and Taigman]{make-a-video}
Uriel Singer, Adam Polyak, Thomas Hayes, Xi Yin, Jie An, Songyang Zhang, Qiyuan Hu, Harry Yang, Oron Ashual, Oran Gafni, Devi Parikh, Sonal Gupta, and Yaniv Taigman.
\newblock Make-a-video: Text-to-video generation without text-video data.
\newblock In \emph{The Eleventh International Conference on Learning Representations, {ICLR} 2023, Kigali, Rwanda, May 1-5, 2023}. OpenReview.net, 2023{\natexlab{a}}.

\bibitem[Singer et~al.(2023{\natexlab{b}})Singer, Sheynin, Polyak, Ashual, Makarov, Kokkinos, Goyal, Vedaldi, Parikh, Johnson, and Taigman]{make-a-video4d}
Uriel Singer, Shelly Sheynin, Adam Polyak, Oron Ashual, Iurii Makarov, Filippos Kokkinos, Naman Goyal, Andrea Vedaldi, Devi Parikh, Justin Johnson, and Yaniv Taigman.
\newblock Text-to-4d dynamic scene generation.
\newblock In \emph{International Conference on Machine Learning, {ICML} 2023, 23-29 July 2023, Honolulu, Hawaii, {USA}}, pages 31915--31929. {PMLR}, 2023{\natexlab{b}}.

\bibitem[Song et~al.(2020)Song, Meng, and Ermon]{ddim}
Jiaming Song, Chenlin Meng, and Stefano Ermon.
\newblock Denoising diffusion implicit models.
\newblock \emph{arXiv preprint arXiv:2010.02502}, 2020.

\bibitem[Stearns et~al.(2024)Stearns, Harley, Uy, Dubost, Tombari, Wetzstein, and Guibas]{dymarbles}
Colton Stearns, Adam~W. Harley, Mikaela~Angelina Uy, Florian Dubost, Federico Tombari, Gordon Wetzstein, and Leonidas~J. Guibas.
\newblock Dynamic gaussian marbles for novel view synthesis of casual monocular videos.
\newblock In \emph{{SIGGRAPH} Asia 2024 Conference Papers, {SA} 2024, Tokyo, Japan, December 3-6, 2024}, pages 30:1--30:11. {ACM}, 2024.

\bibitem[Sun et~al.(2024{\natexlab{a}})Sun, Guo, Wan, Yan, Yin, Zhou, Liao, and Li]{eg4d}
Qi Sun, Zhiyang Guo, Ziyu Wan, Jing~Nathan Yan, Shengming Yin, Wengang Zhou, Jing Liao, and Houqiang Li.
\newblock {EG4D:} explicit generation of 4d object without score distillation.
\newblock \emph{CoRR}, abs/2405.18132, 2024{\natexlab{a}}.

\bibitem[Sun et~al.(2024{\natexlab{b}})Sun, Chen, Liu, Chen, Duan, Zhang, and Wang]{dimensionX}
Wenqiang Sun, Shuo Chen, Fangfu Liu, Zilong Chen, Yueqi Duan, Jun Zhang, and Yikai Wang.
\newblock Dimensionx: Create any 3d and 4d scenes from a single image with controllable video diffusion.
\newblock \emph{CoRR}, abs/2411.04928, 2024{\natexlab{b}}.

\bibitem[Team(2024)]{kling}
KLING~AI Team.
\newblock Kling image-to-video model, 2024.

\bibitem[Teed and Deng(2021)]{raft}
Zachary Teed and Jia Deng.
\newblock {RAFT:} recurrent all-pairs field transforms for optical flow (extended abstract).
\newblock In \emph{Proceedings of the Thirtieth International Joint Conference on Artificial Intelligence, {IJCAI} 2021, Virtual Event / Montreal, Canada, 19-27 August 2021}, pages 4839--4843. ijcai.org, 2021.

\bibitem[Tulyakov et~al.(2018)Tulyakov, Liu, Yang, and Kautz]{tulyakov2018mocogan}
Sergey Tulyakov, Ming-Yu Liu, Xiaodong Yang, and Jan Kautz.
\newblock Mocogan: Decomposing motion and content for video generation.
\newblock In \emph{Proceedings of the IEEE conference on computer vision and pattern recognition}, pages 1526--1535, 2018.

\bibitem[Vondrick et~al.(2016)Vondrick, Pirsiavash, and Torralba]{DBLP:conf/nips/VondrickPT16}
Carl Vondrick, Hamed Pirsiavash, and Antonio Torralba.
\newblock Generating videos with scene dynamics.
\newblock In \emph{Advances in Neural Information Processing Systems 29: Annual Conference on Neural Information Processing Systems 2016, December 5-10, 2016, Barcelona, Spain}, pages 613--621, 2016.

\bibitem[Wang et~al.(2024{\natexlab{a}})Wang, Liu, Liu, Wang, Dong, and Yang]{vistadream}
Haiping Wang, Yuan Liu, Ziwei Liu, Wenping Wang, Zhen Dong, and Bisheng Yang.
\newblock Vistadream: Sampling multiview consistent images for single-view scene reconstruction.
\newblock \emph{arXiv preprint arXiv:2410.16892}, 2024{\natexlab{a}}.

\bibitem[Wang et~al.(2024{\natexlab{b}})Wang, Leroy, Cabon, Chidlovskii, and Revaud]{dust3r}
Shuzhe Wang, Vincent Leroy, Yohann Cabon, Boris Chidlovskii, and Jerome Revaud.
\newblock Dust3r: Geometric 3d vision made easy.
\newblock In \emph{Proceedings of the IEEE/CVF Conference on Computer Vision and Pattern Recognition}, pages 20697--20709, 2024{\natexlab{b}}.

\bibitem[Wang et~al.(2020)Wang, Bilinski, Br{\'{e}}mond, and Dantcheva]{DBLP:conf/wacv/WangBBD20}
Yaohui Wang, Piotr Bilinski, Fran{\c{c}}ois Br{\'{e}}mond, and Antitza Dantcheva.
\newblock Imaginator: Conditional spatio-temporal {GAN} for video generation.
\newblock In \emph{{IEEE} Winter Conference on Applications of Computer Vision, {WACV} 2020, Snowmass Village, CO, USA, March 1-5, 2020}, pages 1149--1158. {IEEE}, 2020.

\bibitem[Wang et~al.(2024{\natexlab{c}})Wang, Chen, Ma, Zhou, Huang, Wang, Yang, He, Yu, Yang, et~al.]{Lavie}
Yaohui Wang, Xinyuan Chen, Xin Ma, Shangchen Zhou, Ziqi Huang, Yi Wang, Ceyuan Yang, Yinan He, Jiashuo Yu, Peiqing Yang, et~al.
\newblock Lavie: High-quality video generation with cascaded latent diffusion models.
\newblock \emph{International Journal of Computer Vision}, pages 1--20, 2024{\natexlab{c}}.

\bibitem[Wang et~al.(2024{\natexlab{d}})Wang, He, Li, Li, Yu, Ma, Li, Chen, Chen, Wang, Luo, Liu, Wang, Wang, and Qiao]{viclip}
Yi Wang, Yinan He, Yizhuo Li, Kunchang Li, Jiashuo Yu, Xin Ma, Xinhao Li, Guo Chen, Xinyuan Chen, Yaohui Wang, Ping Luo, Ziwei Liu, Yali Wang, Limin Wang, and Yu Qiao.
\newblock Internvid: {A} large-scale video-text dataset for multimodal understanding and generation.
\newblock In \emph{The Twelfth International Conference on Learning Representations, {ICLR} 2024, Vienna, Austria, May 7-11, 2024}. OpenReview.net, 2024{\natexlab{d}}.

\bibitem[Wang et~al.(2023)Wang, Lu, Wang, Bao, Li, Su, and Zhu]{prolificdreamer}
Zhengyi Wang, Cheng Lu, Yikai Wang, Fan Bao, Chongxuan Li, Hang Su, and Jun Zhu.
\newblock Prolificdreamer: High-fidelity and diverse text-to-3d generation with variational score distillation.
\newblock \emph{Advances in Neural Information Processing Systems}, 36:\penalty0 8406--8441, 2023.

\bibitem[Wei et~al.(2024)Wei, Zhou, Sun, and Zhang]{wei2024adversarial}
Min Wei, Jingkai Zhou, Junyao Sun, and Xuesong Zhang.
\newblock Adversarial score distillation: when score distillation meets gan.
\newblock In \emph{Proceedings of the IEEE/CVF Conference on Computer Vision and Pattern Recognition}, pages 8131--8141, 2024.

\bibitem[Wu et~al.(2024{\natexlab{a}})Wu, Yi, Fang, Xie, Zhang, Wei, Liu, Tian, and Wang]{4dgs}
Guanjun Wu, Taoran Yi, Jiemin Fang, Lingxi Xie, Xiaopeng Zhang, Wei Wei, Wenyu Liu, Qi Tian, and Xinggang Wang.
\newblock 4d gaussian splatting for real-time dynamic scene rendering.
\newblock In \emph{Proceedings of the IEEE/CVF conference on computer vision and pattern recognition}, pages 20310--20320, 2024{\natexlab{a}}.

\bibitem[Wu et~al.(2024{\natexlab{b}})Wu, Gao, Poole, Trevithick, Zheng, Barron, and Holynski]{cat4d}
Rundi Wu, Ruiqi Gao, Ben Poole, Alex Trevithick, Changxi Zheng, Jonathan~T Barron, and Aleksander Holynski.
\newblock Cat4d: Create anything in 4d with multi-view video diffusion models.
\newblock \emph{arXiv preprint arXiv:2411.18613}, 2024{\natexlab{b}}.

\bibitem[Yang et~al.(2024{\natexlab{a}})Yang, Gao, Zhou, Jiao, Zhang, and Jin]{DBLP:conf/cvpr/YangGZJ0024}
Ziyi Yang, Xinyu Gao, Wen Zhou, Shaohui Jiao, Yuqing Zhang, and Xiaogang Jin.
\newblock Deformable 3d gaussians for high-fidelity monocular dynamic scene reconstruction.
\newblock In \emph{{IEEE/CVF} Conference on Computer Vision and Pattern Recognition, {CVPR} 2024, Seattle, WA, USA, June 16-22, 2024}, pages 20331--20341. {IEEE}, 2024{\natexlab{a}}.

\bibitem[Yang et~al.(2024{\natexlab{b}})Yang, Gao, Zhou, Jiao, Zhang, and Jin]{yang2024deformable}
Ziyi Yang, Xinyu Gao, Wen Zhou, Shaohui Jiao, Yuqing Zhang, and Xiaogang Jin.
\newblock Deformable 3d gaussians for high-fidelity monocular dynamic scene reconstruction.
\newblock In \emph{Proceedings of the IEEE/CVF conference on computer vision and pattern recognition}, pages 20331--20341, 2024{\natexlab{b}}.

\bibitem[Yin et~al.(2023)Yin, Xu, Wang, Zhao, and Wei]{4dgen}
Yuyang Yin, Dejia Xu, Zhangyang Wang, Yao Zhao, and Yunchao Wei.
\newblock 4dgen: Grounded 4d content generation with spatial-temporal consistency.
\newblock \emph{CoRR}, abs/2312.17225, 2023.

\bibitem[Yu et~al.(2024{\natexlab{a}})Yu, Wang, Zhuang, Menapace, Siarohin, Cao, Jeni, Tulyakov, and Lee]{4real}
Heng Yu, Chaoyang Wang, Peiye Zhuang, Willi Menapace, Aliaksandr Siarohin, Junli Cao, L{\'{a}}szl{\'{o}}~A. Jeni, Sergey Tulyakov, and Hsin{-}Ying Lee.
\newblock 4real: Towards photorealistic 4d scene generation via video diffusion models.
\newblock In \emph{Advances in Neural Information Processing Systems 38: Annual Conference on Neural Information Processing Systems 2024, NeurIPS 2024, Vancouver, BC, Canada, December 10 - 15, 2024}, 2024{\natexlab{a}}.

\bibitem[Yu et~al.(2024{\natexlab{b}})Yu, Xing, Yuan, Hu, Li, Huang, Gao, Wong, Shan, and Tian]{viewcrafter}
Wangbo Yu, Jinbo Xing, Li Yuan, Wenbo Hu, Xiaoyu Li, Zhipeng Huang, Xiangjun Gao, Tien-Tsin Wong, Ying Shan, and Yonghong Tian.
\newblock Viewcrafter: Taming video diffusion models for high-fidelity novel view synthesis.
\newblock \emph{arXiv preprint arXiv:2409.02048}, 2024{\natexlab{b}}.

\bibitem[Zhang et~al.(2024)Zhang, Herrmann, Hur, Jampani, Darrell, Cole, Sun, and Yang]{monst3r}
Junyi Zhang, Charles Herrmann, Junhwa Hur, Varun Jampani, Trevor Darrell, Forrester Cole, Deqing Sun, and Ming-Hsuan Yang.
\newblock Monst3r: A simple approach for estimating geometry in the presence of motion.
\newblock \emph{arXiv preprint arXiv:2410.03825}, 2024.

\bibitem[Zhang et~al.(2018)Zhang, Isola, Efros, Shechtman, and Wang]{lpips}
Richard Zhang, Phillip Isola, Alexei~A Efros, Eli Shechtman, and Oliver Wang.
\newblock The unreasonable effectiveness of deep features as a perceptual metric.
\newblock In \emph{Proceedings of the IEEE conference on computer vision and pattern recognition}, pages 586--595, 2018.

\bibitem[Zhao et~al.(2023)Zhao, Yan, Xie, Hong, Li, and Lee]{animate124}
Yuyang Zhao, Zhiwen Yan, Enze Xie, Lanqing Hong, Zhenguo Li, and Gim~Hee Lee.
\newblock Animate124: Animating one image to 4d dynamic scene.
\newblock \emph{CoRR}, abs/2311.14603, 2023.

\bibitem[Zhao et~al.(2024)Zhao, Lin, Lin, Yan, Li, Yang, Wang, Lee, and Wang]{genXD}
Yuyang Zhao, Chung{-}Ching Lin, Kevin Lin, Zhiwen Yan, Linjie Li, Zhengyuan Yang, Jianfeng Wang, Gim~Hee Lee, and Lijuan Wang.
\newblock Genxd: Generating any 3d and 4d scenes.
\newblock \emph{CoRR}, abs/2411.02319, 2024.

\bibitem[Zheng et~al.(2024)Zheng, Li, Nagano, Liu, Hilliges, and De~Mello]{dream-in-4d}
Yufeng Zheng, Xueting Li, Koki Nagano, Sifei Liu, Otmar Hilliges, and Shalini De~Mello.
\newblock A unified approach for text-and image-guided 4d scene generation.
\newblock In \emph{Proceedings of the IEEE/CVF Conference on Computer Vision and Pattern Recognition}, pages 7300--7309, 2024.

\end{thebibliography}
}

\clearpage
\appendix
\clearpage
\setcounter{figure}{0}
\setcounter{section}{0}
\setcounter{table}{0}
\renewcommand{\thefigure}{{\Alph{figure}}}
\renewcommand{\thetable}{{\Alph{table}}}
\renewcommand{\thesection}{{\Alph{section}}}
\maketitlesupplementary



\section{More Implementation Details}
\label{supp:more-implementation-details}

\noindent\textbf{4D-GS Network.} 4D Gaussian Splatting (4D-GS)~\cite{4dgs} lies in extending static 3D Gaussian primitives~\cite{3dgs} to dynamically model temporal-spatial scenes. In 3D-GS~\cite{3dgs}, a scene is represented by a set of anisotropic Gaussians $\mathcal{G} = \{g_i \}_{i=1}^N$, where each Gaussian $g_i$ is parameterized by its position $\mu_i \in \mathbb{R}^3$, rotation (quaternion $q_i \in \mathbb{R}^4$), scale $s_i \in \mathbb{R}^3$, and opacity $\alpha_i \in [0,1]$. The covariance matrix $\Sigma_i$ is derived from $q_i$ and $s_i$, enabling differentiable rendering via splatting.

To model 4D dynamics, each Gaussian is further augmented with time-varying parameters. For temporal coherence, we parameterize the trajectory of $g_i$ over time $t$ through a deformation function $\Delta: \mathbb{R}^4 \to \mathbb{R}^9$:
\begin{equation}
    [\Delta \mu_i(t), \Delta q_i(t), \Delta s_i(t)] = \Delta(\mu_i, q_i, s_i, t),
\end{equation}
where $\Delta$ can be implemented via MLPs or explicit keyframe interpolation. The interpolated Gaussian $g_i(t)$ at time $t$ is then rendered following the 3D-GS rendering pipeline, but with all parameters conditioned on $t$.
Optimization typically requires multi-view RGB videos with camera poses.
While achieving real-time dynamic rendering (30+ FPS), 4D-GS depends heavily on consistent multi-view video supervision.

\noindent\textbf{Training Setup.} We adopt the 4D representation proposed in~\cite{4dgs}. Our hyperparameter settings mainly follow those in~\cite{4dgs}. The learning rate is initialized at \( 1.6 \times 10^{-3} \) and gradually decays to \( 1.6 \times 10^{-4} \) by the end of training. The Gaussian deformation decoder, implemented as a tiny MLP, starts with a learning rate of \( 1.6 \times 10^{-4} \), which is reduced to \( 1.6 \times 10^{-5} \) over time. The training batch size is set to $1$. During the coarse stage, we train for $9$k iterations, followed by an additional $1$k iterations in the fine stage. The $\lambda$ used in the fine-stage loss is $0.1$.
In modulation-based refinement, \( \bar{T} \) is set to $5$ to improve efficiency, and \( w_i \) linearly decreases from 0.5 to 0. Viewcrafter~\cite{viewcrafter} uses its default denoising steps, which is 50. The guidance scale $s$ used in CFG is the default value $7.5$. For multi-view image generation at the first timestamp \( t=1 \), we use adaptive CFG. For \( t>1 \), CFG is disabled because the reference information from the multi-view generation at \( t=1 \) has already been introduced into the missing regions. All experiments are conducted on a single NVIDIA A100 (40GB) GPU.

\section{Details of User Study}
\label{supp:details-of-user-study}

\begin{figure}[!t]
  \centering
    \includegraphics[width=0.99\linewidth]{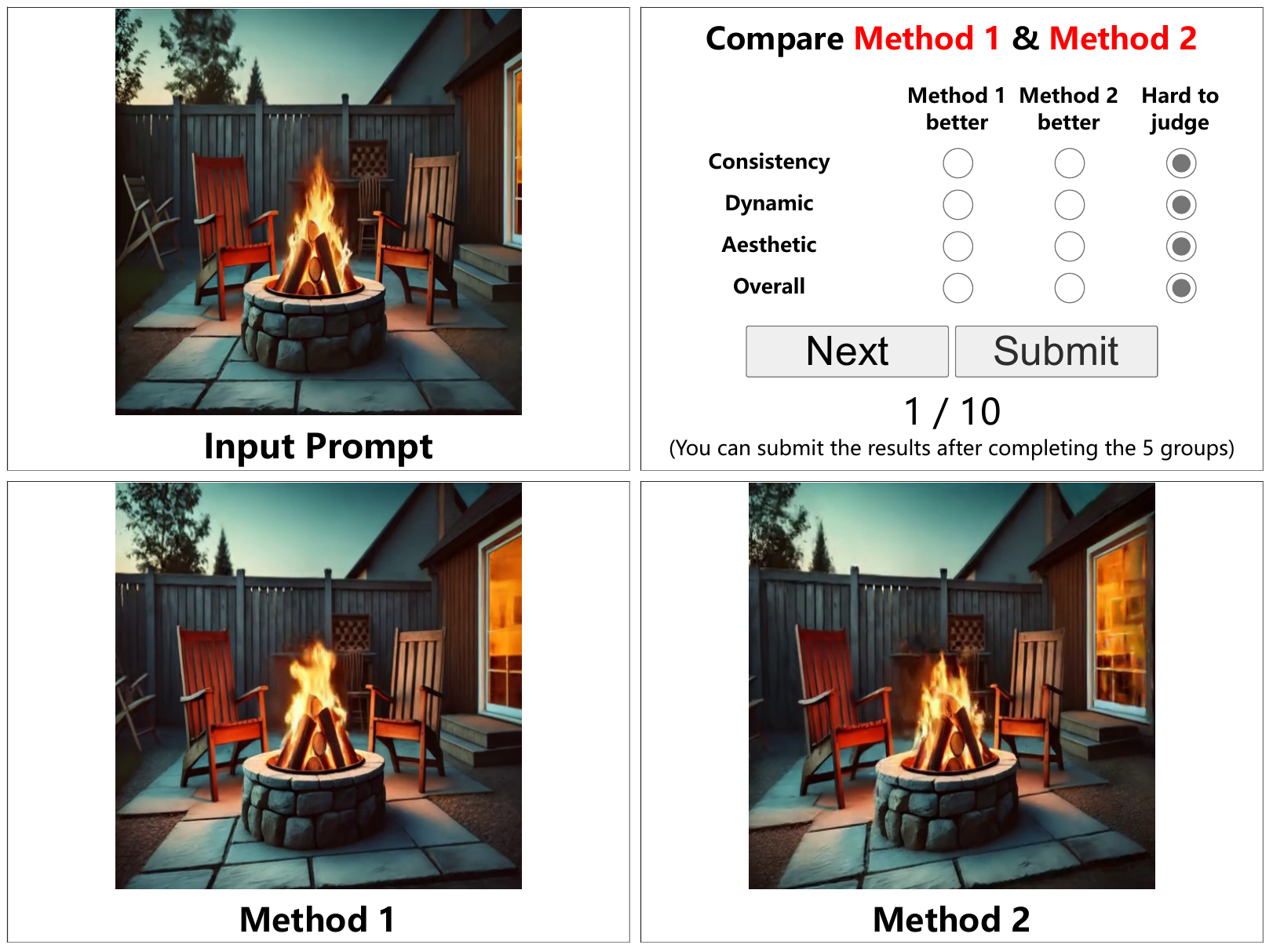}
    \caption{\textbf{The web interface of our user studies.} The input prompt can be either a single image or a short text. 
    }
    \label{fig:userstudy-ui}
\end{figure}

\noindent \textbf{User Study I: Comparison with Other Methods.} We conducted the first user study to compare our method with other existing methods. Since the source codes of these methods were not publicly available, we compared our method with the videos provided on their respective project pages. A total of $32$ pairs of videos were used in this study. Each pair was generated from the same input images or text prompts to ensure a fair comparison. The methods included in this study were 4Real~\cite{4real}, 4Dfy~\cite{4dfy}, Dream-in-4D~\cite{dream-in-4d}, DimensionX~\cite{dimensionX}, GenXD~\cite{genXD}, and Animate124~\cite{animate124}. The user study was conducted online, and a screenshot of the interface is shown in~\cref{fig:userstudy-ui}. Participants were asked to evaluate the generated videos based on four criteria: Consistency, Dynamic, Aesthetic, and Overall. For each pair of videos, they were required to select which method performed better for each criterion. They could skip to the next example without selecting if they found it difficult to judge. The user study was conducted anonymously, and no personally identifiable data were collected.

\noindent \textbf{User Study II: Ablation Study.} The second user study evaluated the impact of our method's different components through an ablation experiment. The components included in this study were \textit{Monst3R}, \textit{Adaptive CFG}, \textit{Point Cloud Guided Denoising}, \textit{Reference Latent Replacement}, \textit{Reference Latent Replacement}, \textit{Coarse-to-fine optimization}, and \textit{Modulation-based Refinement}. For each component, we randomly sampled $10$ different scenes and generated video pairs using the full version of our method and a variant with the specific component removed or modified. Participants were asked to evaluate the generated video pairs based on the same four criteria as in User Study I: consistency, aesthetics, motion dynamic, and overall quality. The ablation study was also conducted anonymously, without collecting any personally identifiable data.

\section{Details of VBench Metrics}
\label{supp:details-of-vbench-metrics}
To evaluate the quality of multi-view videos rendered from 4D representations, we report common VBench~\cite{vbench} metrics: Consistency (average for subject/background), Dynamic Degree, Aesthetic Score, and Text Alignment (only for text-to-4D). 

\noindent \textbf{Subject / Background Consistency.} To evaluate the consistency of both subjects (\eg, a person, car, or cat) and background scenes in the video, VBench uses DINO~\cite{dino} and CLIP~\cite{clip} feature similarities across frames. DINO captures subject consistency by comparing frame embeddings, while CLIP assesses background stability. Together, they provide a comprehensive measure of consistency.

\noindent \textbf{Dynamic Degree.} Since a static video can score well in the aforementioned consistency metrics, it is important to evaluate the degree of dynamics (\ie, whether it contains large motions). To this end, the Dynamic Degree metric uses RAFT~\cite{raft} to estimate the degree of dynamics in synthesized videos.  Specifically, this metric takes the average of the largest $5\%$ optical flows (considering the movement of small objects in the video). This approach ensures that minor movements (\eg, small objects or slight camera shakes) do not disproportionately influence the overall dynamic assessment.

\noindent \textbf{Aesthetic Score.} We evaluate the artistic and beauty value perceived by humans towards each video frame using the LAION Aesthetic Predictor~\cite{LAION}. This predictor is a linear model built on top of CLIP embeddings, trained to assess the aesthetic quality of images on a scale from $1$ to $10$. It reflects various aesthetic aspects, including the layout, richness and harmony of colors, photo-realism, naturalness, and overall artistic quality of the video frames. The Aesthetic Score metric obtains a normalized aesthetic score by applying this predictor to each frame.

\noindent \textbf{Text Alignment.} This metric uses overall video-text consistency computed by ViCLIP~\cite{viclip} on general text prompts as an aiding metric to reflect text semantics consistency. ViCLIP is a video-text contrastive learning model that leverages a large-scale video-text dataset to learn robust and transferable representations. 

\section{Runtime Analysis}  
\label{supp:runtime-analysis}
The runtime comparison is shown in~\Cref{tab:runtime}.  
We compare our approach with object-level methods~\cite{4dfy} and the text-to-4D scene generation method~\cite{4real}.  
Since \cite{dimensionX} and \cite{genXD} have not reported runtime details (including feed-forward inference time and 4D representation optimization time) or released their code, they are excluded from the comparison.  
Notably, compared to previous methods, our approach supports higher resolutions while efficiently handling more frames and viewpoints, achieving the fastest optimization.
Our total runtime is composed of three main steps: running MonST3R~($1$ min), generating multi-view videos with ViewCrafter~($25$ min), and optimizing 4D-GS~($35$ min).

\begin{table}[!t]
    \small
  \centering
  \setlength{\tabcolsep}{8pt}
  \begin{tabular}{c|cccc}
    \toprule
     Method & Time & Resolution & Frames & Views\\ 
     \midrule
     4Dfy~\cite{4dfy}   & 10h+  & 256$\times$256    & -     & -     \\
     4Real~\cite{4real} & 1.5h  & 256$\times$144    & 8     & 16    \\
     Ours               & 1h    & 1024$\times$576   & 16    & 25    \\
     \bottomrule
    \end{tabular}
    \caption{\textbf{Comparison of runtime with other methods.} Frames and Views represent the number of video frames and the number of viewpoints, respectively. The running time of Structure from Motion (SfM), such as colmap~\cite{colmap}, is not included due to significant variations across different scenes.
    }
  \label{tab:runtime}
\end{table}

\begin{figure}[!t]
  \centering
    \includegraphics[width=0.99\linewidth]{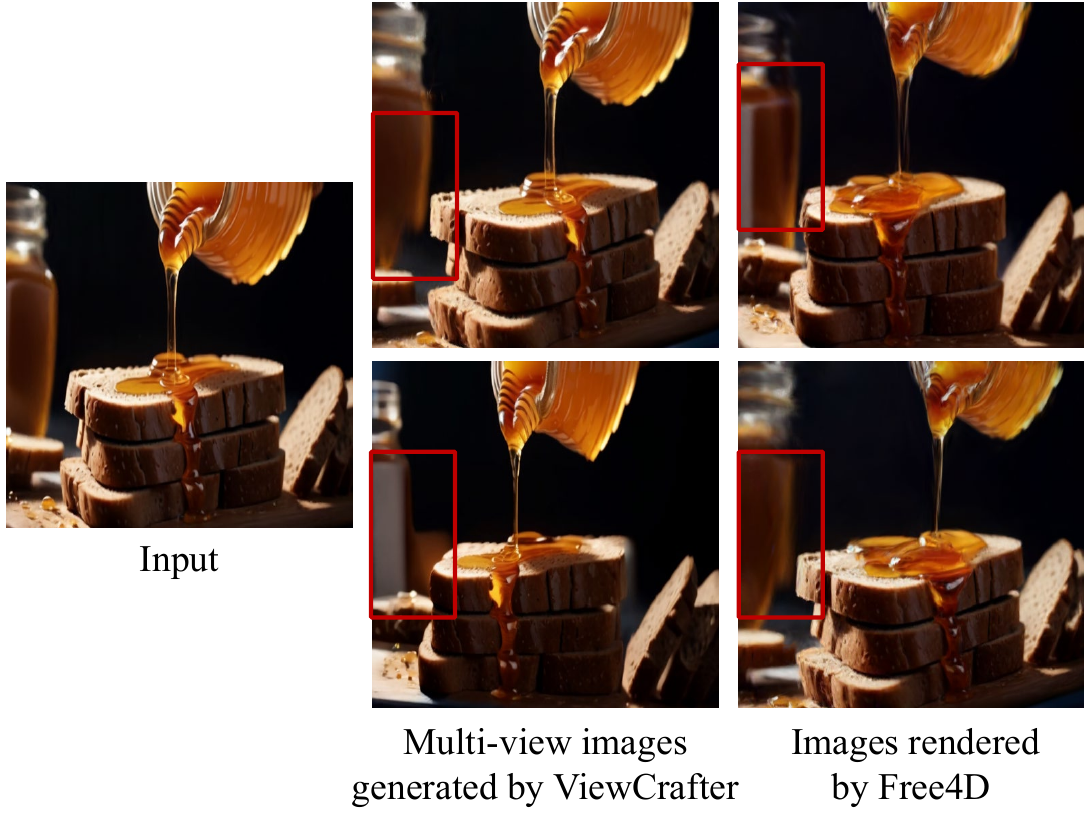}
    \caption{\textbf{Failure Case.} ViewCrafter~\cite{viewcrafter} struggles with blurred or defocused regions, leading to distortions that propagate into the 4DGS-rendered results.}
    \label{fig:failure-case}
    \vspace{-10pt}
\end{figure}

\section{Limitations and Future Work} 
\label{supp:limitations-and-future-work}
\noindent \textbf{Limitations.} Since our method primarily relies on the prior from ViewCrafter~\cite{viewcrafter} to generate consistent multi-view videos, it also inherits some of its limitations.
Firstly, it struggles to synthesize novel views with large view ranges from limited 3D clues, such as generating a front view from only a back view.  
Additionally, since ViewCrafter depends on accurate point cloud geometry, it has difficulty handling severely blurred or defocused regions, which hinder depth estimation, as shown in~\Cref{fig:failure-case}.

\noindent \textbf{Future Work.} We recognize that the accuracy of MonST3R~\cite{monst3r}'s estimation of dynamic videos is crucial. We observed that Dust3R~\cite{dust3r} demonstrates better robustness than MonST3R in some static scenes. Therefore, a potential approach is to use Dust3R to estimate the geometry of the first frame, and employ optical flow to link different views during the subsequent 4DGS optimization.

\end{document}